\documentclass[lettersize,journal]{IEEEtran}
\usepackage{amsmath,amsfonts}
\usepackage{algorithmic}
\usepackage{algorithm}
\usepackage{array}
\usepackage[caption=false,font=normalsize,labelfont=sf,textfont=sf]{subfig}
\usepackage{textcomp}
\usepackage{url}
\usepackage{verbatim}
\usepackage{graphicx}
\usepackage{amssymb, booktabs}
\usepackage{epsfig}
\usepackage{multirow}
\usepackage{tabu}                     
\usepackage{multicol}                 
\usepackage{multirow}                
\usepackage{float}                    
\usepackage{makecell}                 
\usepackage{booktabs}                 
\usepackage{float}                    
\usepackage{graphicx}
\usepackage{xcolor}
\usepackage{colortbl}
\usepackage{booktabs}
\usepackage{placeins}
\usepackage{subcaption}
\usepackage{fontawesome}
\usepackage{amsmath}
\usepackage[pagebackref=true,breaklinks=true,letterpaper=true,colorlinks,citecolor=blue,linkcolor=blue,bookmarks=false]{hyperref}

\begin{document}

\title{Hazy Pedestrian Trajectory Prediction via Physical Priors and Graph-Mamba}

\author{ 
    Jian Chen$^{1*}$,
    Zhuoran Zheng$^{2*}$,
	Han Hu$^{1}$, \\
	Guijuan Zhang$^{1}$,
	Dianjie Lu$^{1}$, 
	Liang Li$^{3}$,
    Chen Lyu\text{(\faEnvelopeO)}$^{1}$\\
	$^{1}$ Shandong Normal University \\
	$^{2}$ Sun Yat-sen University 
	$^{3}$ Shandong Jiaotong University\\
    {\tt\small bujidui@gmail.com, zhengzhr@mail.sysu.edu.cn, huhan199908@163.com}\\
    {\tt\small \{lvchen, zhangguijuan, ludianjie\}@sdnu.edu.cn, 240050@sdjtu.edu.cn}
    }

\maketitle

\renewcommand{\thefootnote}{} 
\footnotetext{* Equal contribution.}

\begin{abstract}
To address the issues of physical information degradation and ineffective pedestrian interaction modeling in pedestrian trajectory prediction under hazy weather conditions, we propose a deep learning model that combines physical priors of atmospheric scattering with topological modeling of pedestrian relationships.
Specifically, we first construct a differentiable atmospheric scattering model that decouples haze concentration from light degradation through a network with physical parameter estimation, enabling the learning of haze-mitigated feature representations. 
Second, we design an adaptive scanning state space model for feature extraction. Our adaptive Mamba variant achieves a 78\% inference speed increase over native Mamba while preserving long-range dependency modeling.

Finally, to efficiently model pedestrian relationships, we develop a heterogeneous graph attention network, using graph matrices to model multi-granularity interactions between pedestrians and groups, combined with a spatio-temporal fusion module to capture the collaborative evolution patterns of pedestrian movements. 
Furthermore, we constructed a new pedestrian trajectory prediction dataset based on ETH/UCY to evaluate the effectiveness of the proposed method. Experiments show that our method reduces the minADE / minFDE metrics by 37.2\% and 41.5\%, respectively, compared to the SOTA models in dense haze scenarios (visibility $\textless$ 30m), providing a new modeling paradigm for reliable perception in intelligent transportation systems in adverse environments.
\end{abstract}


\section{Introduction}
\begin{figure}[ht]
\centering
\subfloat[]{\includegraphics[width=0.50\columnwidth]{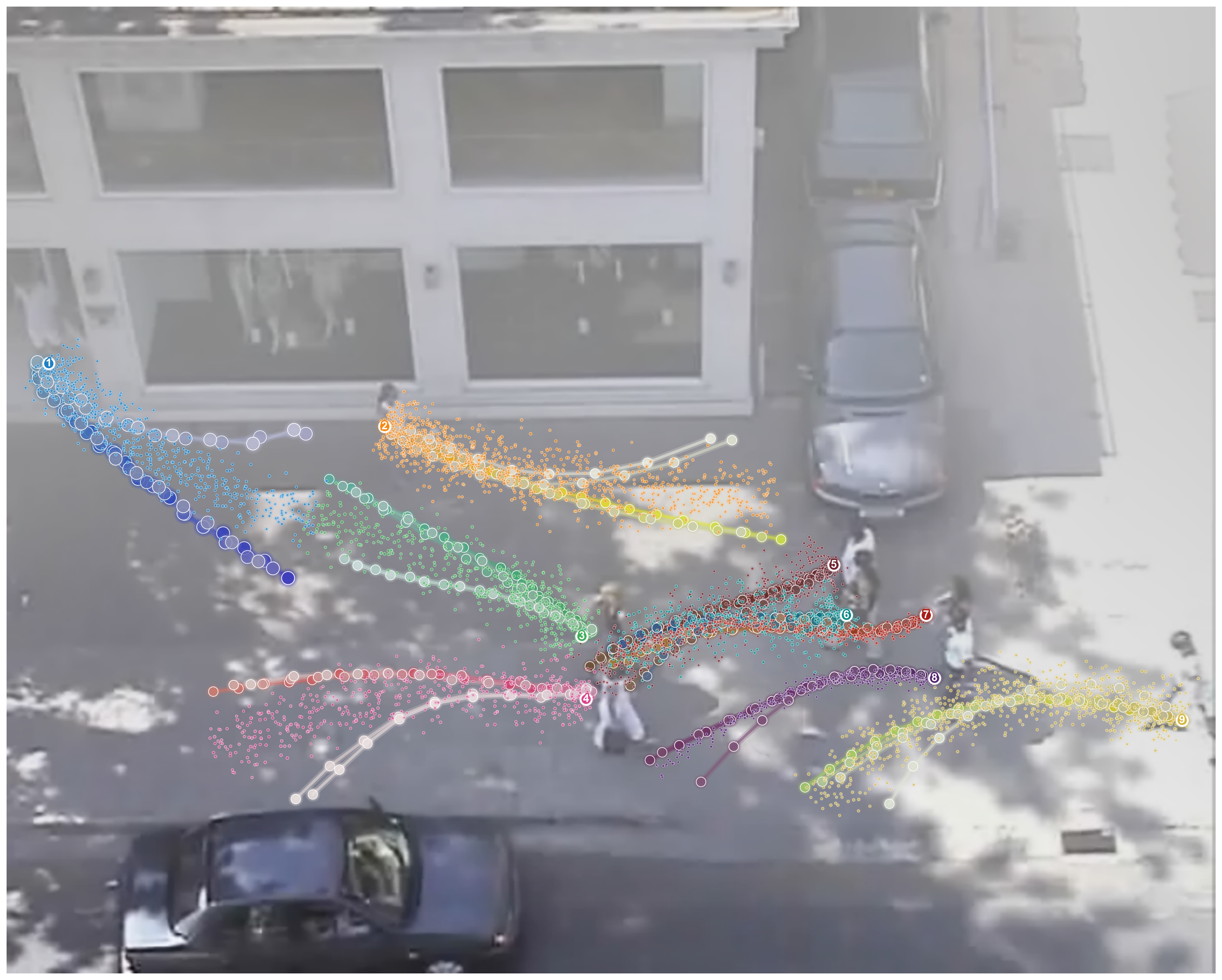}%
\label{fig_first_case}}
\hfil
\subfloat[]{\includegraphics[width=0.50\columnwidth]{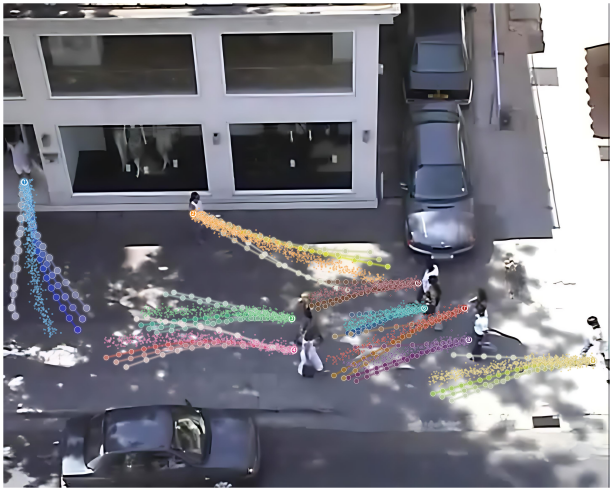}%
\label{fig_second_case}}
\caption{Optical scattering in hazy environments degrades the quality of visual perception system features, while also altering pedestrian behavior patterns, such as reduced speed, tighter group formations, and contracted social interaction ranges. Therefore, existing pedestrian trajectory prediction methods struggle to achieve accurate predictions under hazy weather conditions.
(a) Pedestrian trajectory prediction in haze. (b) Pedestrian trajectory prediction in non-hazy conditions.}
\label{figs_1}
\vspace{-4mm}
\end{figure}

\IEEEPARstart{W}{ith} the rapid development of intelligent transportation and urban management systems, pedestrian trajectory prediction technology has become a key to ensuring traffic safety and improving urban operational efficiency. 
However, in practical application environments, low visibility weather conditions such as heavy haze occur frequently, as shown in Figure~\ref{figs_1}, posing a dual challenge to pedestrian trajectory prediction systems in such hazy environments.

So far, pedestrian trajectory prediction technology has experienced a development process from traditional physical models to deep learning methods. Early studies have mainly simulated pedestrians as stressed particles, but have struggled to capture complex intentions and social interactions~\cite{helbing1995social}. In contrast, based on its powerful feature learning capabilities, deep learning has made significant progress in capturing complex pedestrian behavior patterns and social interactions, and can better handle large amounts of pedestrian trajectory data and mine potential laws \cite{9762760}. However, existing deep learning methods have the following obstacles in the pedestrian trajectory prediction in adverse weather:
\textbf{i)} Ignoring the impact of physical degradation caused by adverse weather on visual features.
\textbf{ii)} Excessive reliance on self-attention-based architectures (e.g., Transformer) leads to high computational complexity when processing a long sequence of tokens.
\textbf{iii)} The lack of an effective pedestrian group interaction model in hazy days cannot accurately capture the collaborative decision-making and spatial distribution changes among groups in adverse weather.
These challenges severely restrict the robustness and generalization ability of pedestrian trajectory prediction techniques in practical, complex scenarios.


In response to the aforementioned three key challenges, we model the dehazing of visual degradation phenomena in hazy environments based on atmospheric scattering theory (Eq. \eqref{deqn_ex1a1}). 
According to the Koschmieder scattering model theory of \cite{9577761}, image or video information can be characterized as a linear superposition of the reflected light from an object and the atmospheric scattered light, and its intensity is negatively correlated with the scattering coefficient. Based on this theory, we first built a Physical Parameter Estimation Network (PPEN). The network adopts the U-Net encoder-decoder architecture, realizes the estimation from hazy image $\mathbf{I}(\textbf{x})$ to the pixel-level depth map $d(\textbf{x})$, and predicts the global atmospheric illumination value ${A}$ through the global feature pooling operation in conjunction with the multilayer perceptron (MLP). In addition, PPEN realizes the accurate estimation of scattering coefficients $\beta$ for a hazy scene by fusing the global features of the haze scene and the local features of the pedestrian clusters. By using the network-estimated physical parameters, we attempt to reconstruct a clear image $\mathbf{I}_0(\textbf{x})$, and achieve end-to-end learning through the optimization of haze-invariant features in \cite{lee2022fifo,ahamed2024timemachine}, laying a theoretical foundation for the prediction of pedestrian trajectories in hazy scenes.

\begin{equation}
\label{deqn_ex1a1}
\mathbf{I}(\textbf{x}) = \mathbf{I}_0(\textbf{x}) \cdot e^{-\beta \: d(\textbf{x})} + A \cdot \left( 1 - e^{-\beta \: d(\textbf{x})} \right).
\end{equation}

Second, to alleviate the high computational complexity of long-range sequence modeling, we introduce the state space model (Mamba) \cite{huang2025trajectory} as the core of temporal modeling and innovatively construct a selective scanning mechanism to efficiently model long-range context with linear complexity. This significantly improves efficiency compared to traditional Transformer and native Mamba (inference speed increased by 78\%). 
Finally, to address the issue of insufficient adaptability to adverse environments in pedestrian social interaction modeling, we introduce graph neural networks to build multi-granularity interaction relationships. By hierarchically aggregating mutual influences among pedestrians and using dynamic graph relationships to adaptively adjust interaction intensity, we effectively solve problems such as reduced social perception range and collapsed interaction patterns in hazy weather scenarios.

Our main contributions are summarized as follows:
\begin{enumerate}
\item{We propose a joint atmospheric scattering physics model and deep learning haze pedestrian trajectory prediction framework, realize end-to-end training from physical perception to behavior prediction, innovatively introduce the Mamba for efficient temporal modeling, and solve the computational complexity bottleneck of the traditional attention mechanism.}
\item{We design a multi-granular social relationship representation framework based on graph neural network, accurately model complex social dynamics among pedestrians using dynamic graphs, combine the spatiotemporal feature fusion mechanism, accurately capture the group behavior patterns, and effectively meet the challenge of the change of the social perception range of pedestrians in the haze scene.}
\item{The first pedestrian trajectory prediction with haze dataset is constructed, and the excellent performance of the proposed method under various hazy intensities is demonstrated through extensive experimental results. The prediction accuracy and robustness of the system under different haze intensities were comprehensively evaluated, providing an evaluation benchmark for future related studies.}
\end{enumerate}

\begin{figure*}[ht]
\centering
\includegraphics[width=\textwidth]{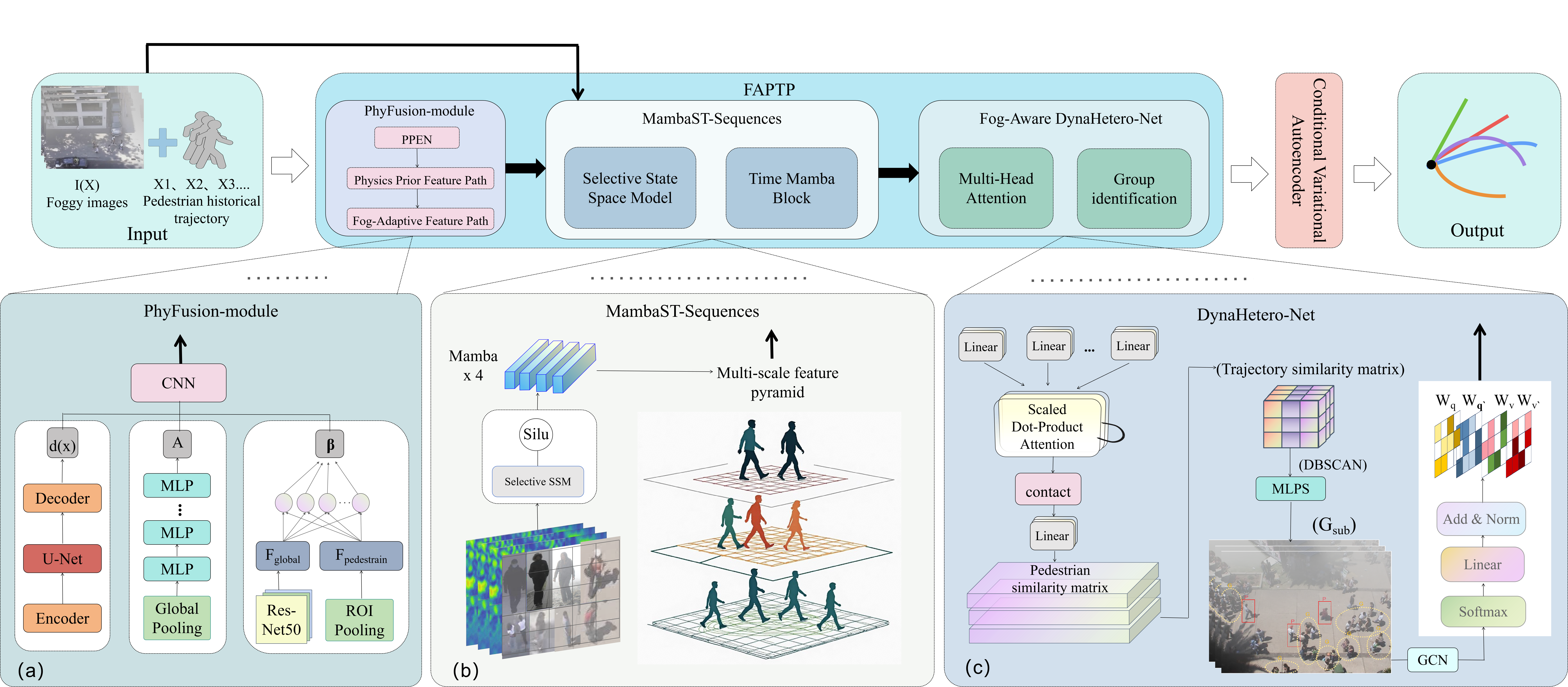} 
\caption{Overview of the proposed FAPTP. Our method includes three key modules: (a) PhyFusion-module achieves the fusion of atmospheric scattering physics prior with deep learning, estimating the scattering parameters through the PPEN network, combined with CNN to extract the haze invariant features; (b) The MambaST-Sequences module receives the scattering parameter features and original information and establishes the selective state space method to extract scenes and pedestrian features with multi-layer feature pyramid; (c) DynaHetero-Net introduces heterogeneous graph attention networks to model social relationships among multi-granular pedestrians, Adaptation to behavioral changes in hazy environments through hierarchical representation and dynamic graph weights. Finally, we predict the trajectory of pedestrians via a conditional variational autoencoder.} 
\label{fig_frameworks}
\end{figure*}

\section{Related Works}
\subsection{Pedestrian Trajectory Prediction}
Pedestrian trajectory prediction techniques have undergone a remarkable development from social attribute modeling to data-driven approaches \cite{10656248,9899358,zhang2021conditional,9899358}. Traditional approaches mainly include the social attribute model \cite{10460178,helbing1995social}, which models pedestrians as individuals with social attributes, but it struggles to capture the intentions and interactions between pedestrians in complex scenes. With the development of deep learning, modern methods such as Social LSTM \cite{7780479,10747507} first modeling pedestrian interaction through the social pooling layer, and Social-GAN \cite{8578338,8953374} using adversarial learning to generate diversified tracks have become mainstream. After that, the focus has gradually shifted to methods for fusion scene understanding, such as the scenario context encoder proposed by Teeti et al. \cite{teeti2025astra} and the graph neural network model of Yu et al. \cite{10498338}, which improve prediction accuracy by integrating environmental information. At the same time, models based on attention mechanisms such as Transformer~\cite{fragkedaki2024pedestrian,9989439,10.1007/978-3-030-58610-2_30,lee2024mart} and diffusion model-based methods \cite{10489838,bae2024singulartrajectory} have also made significant progress. Although these methods perform well under normal weather conditions, they perform ineffectively in low visibility environments and cannot sense special environments. Thus, we fuse the atmospheric scattering physics priors in our deep network, designed to address the challenges of trajectory prediction under anomalous climatic conditions.

\subsection{Mamba for Pedestrian Trajectory Prediction}
Recently, Transformer-based approaches such as Trajectron++ \cite{salzmann2020trajectron++} capture long-range dependencies through a self-attention mechanism, but face memory bottlenecks when dealing with long sequences. The selective state-space model (SSM) proposed by Mamba \cite{ahamed2024timemachine,huang2025trajectory} breaks through this limitation through the scanning mechanism with linear complexity. The mechanism breaks through this limitation. Note that Mamba has been successfully applied to long content generation tasks in the field of natural language processing \cite{zhang2024mamba}, and has demonstrated efficient processing of temporal signals in speech recognition. This study presents a novel integration of the Mamba architecture into pedestrian trajectory prediction, addressing the computational inefficiency of conventional models in processing long sequential trajectories and their limitations in capturing long-term dependencies within complex scenes. By leveraging multi-scale spatio-temporal feature extraction combined with a cross-layer attention mechanism \cite{10012451,ZHANG2023109633,9607725}, the proposed framework significantly enhances both efficiency and predictive accuracy. Second, in complex and changing hazy scenes, the model needs to simultaneously balance computational efficiency and prediction accuracy, which poses a serious challenge to traditional architectures. This motivates us to explore more efficient methods for spatio-temporal sequence modeling.

\subsection{GNN for Pedestrian Trajectory Prediction}
The pedestrian social relationship modeling method can be divided into homogeneous and heterogeneous graphs. Social-GAN uses fully connected graphs to model human interactions, but ignores population structure differences. STGAT \cite{9010834,GUO20241512} introduces a dynamgraph attention mechanism but does not distinguish interaction types. Recent work has begun to explore heterogeneous relationship modeling, such as HGNN \cite{ZHU2023109772}, distinguishing individual-population interactions through hierarchical graph networks, but lacking environmental adaptation. Through the adaptive heterogeneous graph attention network, we dynamically adjust the social relationship modeling, so that the system can adaptively capture the changes of social behavior according to the hazy environment, which significantly improves the rationality of interaction compared with the traditional method.

\subsection{Application of The Atmospheric Scattering Model}
Atmospheric scattering models have demonstrated significant value in various computer vision tasks. The dark channel prior (DCP) proposed by He et al. ~\cite{5206515,7978537,s22145210} became the foundational algorithm for early haze image restoration. In the field of haze target detection, Liu et al. ~\cite{Liu:23} combined scattering models with deep detection networks, significantly enhancing the robustness of pedestrian detection in hazy conditions. Li et al. \cite{10377188} introduced physical scattering constraints in crowd counting tasks under hazy conditions, reducing density estimation bias through adaptive feature enhancement, improving counting accuracy by over 30\% in dense hazy conditions. Chen et al. ~\cite{CHEN2020748} applied scattering models to lane line detection in hazy conditions, enhancing detection accuracy in extreme weather through physically modulated feature representations. Additionally, methods based on scattering models have made significant progress in scene segmentation and depth estimation in haze environments, providing a more reliable perception foundation for safety decisions in autonomous driving systems under adverse weather conditions. To the best of our knowledge, this work represents the first systematic integration of atmospheric scattering physics models into pedestrian trajectory prediction, providing a novel framework for enhancing the robustness of intelligent transportation systems under severe weather conditions.

\section{Methodology}
\subsection{Overview}
As shown in Figure~\ref{fig_frameworks}, our approach consists of three key components, where the input information involves video sequences under hazy weather and extracted pedestrian trajectory data. First, the PhyFusion module extracts features from the input multimodal information to estimate the physical parameters associated with the hazy weather. Next, the MambaST-Sequence module receives the estimated physical parameters and original input information to obtain higher-order features. Meanwhile, DynaHetero-Net integrates spatio-temporal features through a heterogeneous graph attention mechanism and multi-granularity pedestrian interaction modeling to complement the time-awareness capability that MambaST-Sequence does not obtain. Finally, all the feature information is used to predict accurate pedestrian trajectories by a Conditional Variational Auto-Encoder (CVAE) \cite{zhang2021conditional}.

\subsection{PhyFusion Module based on The Physical Prior}
To efficiently estimate the physical parameters of atmospheric scattering for a single image, we develop an end-to-end physical parameter estimation network (PPEN) with three subnetworks. These networks estimate the atmospheric scattering coefficients $\beta$, atmospheric illumination values $A$, and the scene depth map $d(\textbf{x})$ of the image, respectively.

\noindent \textbf{Depth map estimation.}
We use a lightweight CNN to predict the pixel-level depth map $d(\textbf{x})$ from the input image. The network first downsamples the input images through a five-layer encoder to capture multi-scale features, and then recovers the pixel-level depth map with a five-layer decoder. For each layer of the network, we use a $3\times3$ convolutional layer and a ReLU.

\noindent \textbf{Atmospheric estimation.}
We use a global feature pooling and a multilayer perceptron (MLP) to predict atmospheric values $A$. It can be written:
\begin{equation}
\label{eq:A}
A = \mathbf{W}_2 \cdot \mathrm{ReLU}(\mathbf{W}_1 \cdot [\textbf{v}_i, \textbf{p}_i, \exp(-\alpha \cdot d_i)] + \mathbf{b}_1), 
\end{equation}
where $\textbf{v}_i$ is the image pixel value, $\textbf{p}_i$ is the normalized pixel spatial position coordinate, and ${d}_i$ is the estimated depth value. The function contains a global mean pooling and 2 layers of fully connected layers, and the output is the atmospheric light value $A$$\in[0,1]^3$ for the RGB three channels.

\noindent \textbf{Scattering coefficient estimation.}
Combining the global features of the scene and the characteristics of the pedestrian local area, we can predict the scattering coefficient $\beta$ of the current scene:

\begin{equation}
\label{deqn_ex1a}
\beta = \text{g}(\mathbf{F}_{\text{global}}, \mathbf{F}_{\text{pedestrian}}),
\end{equation}
where $\mathbf{F}_{\text{global}}$ is the scene global feature, extracted by ResNet-18 network; $\mathbf{F}_{\text{pedestrian}}$ is the pedestrian area feature, extracted from the pedestrian bounding box by ROI pooling; and g is a 3-layer fully connected network that outputs a single scalar scattering coefficient $\beta$$\in[0,3]$.

\noindent \textbf{Clear image estimation and feature fusion.}
Using the estimated physical parameters, we can obtain a clear image $\mathbf{I}_0(\textbf{x})$ (using Eq.~\ref{deqn_ex1aa}), after which, we develop two complementary feature extraction paths to efficiently provide high-quality features for the downstream feature extractors.  The physical prior feature path  extracts the feature $\mathbf{F}_{\text{phys}}$ from the clear image through the feature extraction network:
\begin{equation}
\label{deqn_ex1aa}
\mathbf{I}_0(\textbf{x}) = \frac{\mathbf{I}(\textbf{x}) - A \cdot \left( 1 - e^{-\beta d(\textbf{x})} \right)}{e^{-\beta d(\textbf{x})}},
\end{equation}

\begin{equation}
\label{deqn_ex1as}
\mathbf{F}_{\text{phys}} = \mathrm{CNN}(\mathbf{I}_0(\textbf{x})),
\end{equation}
where $\mathrm{CNN}$ is a feature extraction network that shares weights with the physical path. The haze-adapted feature path directly extracts the features from the original hazy image $\mathbf{I}$ and adjusts the features through the learnable scattering modulation layer:

\begin{equation}
\mathbf{F}_{\text{adapt}} = \mathrm{CNN}(\mathbf{I}) \odot \big(1 + \gamma \cdot e^{-\beta d_{\text{i}}}\big) + \delta \cdot A \label{eq:wi},
\end{equation}
where $\odot$ represents element-by-element multiplication, $d_{\text{i}}$ is the average depth, representing the depth information of the scene, and $\gamma$ and $\delta$ are learnable parameters, adjusting the scattering coefficients and the effects of atmospheric illumination. Finally, feature fusion is performed by an adaptive attention algorithm:
\begin{align}
\mathbf{F}_{\text{inv}} &= \alpha \cdot \mathbf{F}_{\text{phys}} + (1 - \alpha) \cdot \mathbf{F}_{\text{adapt}}, \label{eq:fvi} 
\end{align}
where $\alpha$ ($[0,1]$) is an adaptive learnable parameter.

\subsection{MambaST-Sequences}
To effectively utilize the feature $\mathbf{F}_{\text{inv}}$, we need an efficient spatio-temporal sequence modeling framework to capture the temporal dependence of pedestrian trajectories.  Here, we design a Mamba-based multiscale spatio-temporal feature extraction framework to efficiently model the multiscale spatio-temporal dependencies of pedestrian trajectories by using multilayer Mamba blocks, cross-layer attention connectivity, and a feature pyramid. This design combines the temporal modeling capability of SSM with spatial feature extraction to form a unified spatio-temporal feature representation framework.
Given a sequence of pedestrian trajectories X= $x_1$, $x_2$, ..., $x_T$, where $x_t\in \mathbb{R} ^{(2+ F)}$ contains two-dimensional coordinates and F-dimensional additional features (velocity, acceleration, direction, etc.), the processing flow of the spatio-temporal Mamba block is as follows:


\begin{equation}
\widehat{\mathbf{X}} = \mathrm{LayerNorm}(\mathbf{X}) \cdot \mathbf{W}_{\text{in}} + \mathbf{b}_{\text{in}},
\end{equation}
where $\mathbf{W}_{\text{in}}\in \mathbb{R} ^{(2+ F) \times D}$ is the projection matrix, $\mathbf{b}_{\text{in}}\in \mathbb{R} ^D$ is the bias term, and $D$ is the model hiding dimension, which is set to 256.
\begin{align}
    & \mathbf{H} \boldsymbol{\Delta}(\textbf{x}) = \mathrm{SSM}(\mathrm{MLPs}(\widehat{\mathbf{X}})), && \label{eq:fvi} \\
    & \mathbf{Y} = \mathbf{C}(\textbf{x}) \cdot \mathbf{H} \odot \mathrm{SiLU}(\widehat{\mathbf{X}} \cdot \mathbf{W}_{\text{gate}}), &&
\end{align}
where $\mathbf{W}_{\text{gate}}\in\mathbb{R}^{D\times D}$ is the gated weight matrix. By mapping the input features to the SSM parameter space, scan updates the states efficiently through the scanning operation, and finally, the output projection $\mathbf{C}(\textbf{x})$ maps the state to the output space $\mathbf{Y}$.
\begin{equation}
\label{deqn_ex1a}
\mathbf{F}_l = \mathrm{LayerNorm}\big(\mathbf{X} + \mathrm{Dropout}(\mathbf{Y} \cdot \mathbf{W}_{\text{out}} + \mathbf{b}_{\text{out}})\big),
\end{equation}
where $\mathbf{W}_{\text{out}}\in \mathbb{R} ^{\mathrm{D\times (2+ F) }}$ is the output projection matrix, and $\mathbf{b}_{\text{out}}\in \mathbb{R} ^{2+ \mathrm{F} }$ is the output bias term. The Dropout is set to 0.1 to provide regularization effects. To enhance information exchange between different time scales, we design cross-layer attentional connections:
\begin{equation}
\label{deqn_ex1a}
\mathbf{F}_{l+1} = \mathbf{F}_l + \alpha_l \cdot \mathrm{Attention}(\mathbf{F}_l, \mathbf{F}_{l-1}).
\end{equation}
Here  $\alpha_l$ is the balance coefficient. In this way, the Mamba blocks in the $l+1$ layer can utilize the relevant spatio-temporal information from the $l- 1$ layer during the processing, thus establishing a more direct connection between different time scales.

After completing the processing of all network layers, we need to fuse the captured multi-scale features into the final representation. Therefore, we introduce an attention-based multi-scale feature pyramid, which fuses the features through the \cite{9373939} feature pyramid mechanism to obtain feature $\mathbf{F}_{\mathrm{multi}}$.

\subsection{DynaHetero-Net}
Multi-scale temporal features extracted through the MambaST-Sequences module capture the temporal dependence of pedestrian motion, but the pedestrian trajectory is also significantly affected by social interactions in complex hazy scenarios. Specifically, we design DynaHetero-Net to deal with the modeling of pedestrian relationships, which can adaptively adjust the social relationships among pedestrians according to the intensity of hazy weather. 
For the pedestrian node $i$, its feature update is implemented through the heterogeneous graph attention mechanism.
To adapt to the hazy environment, we designed a hazy perceptual modulation function $g_{r}$ to dynamically adjust the weight of attention in the network:
\begin{align}
    \alpha_{ij}^{r,\mathrm{haze}} &= \sigma\big(\mathbf{W}_2^r \cdot \mathrm{ReLU}(\mathbf{W}_1^r[\beta, d_{ij}] + \mathbf{b}_1^r) + \mathbf{b}_2^r\big),\label{eq:fvi} 
\end{align}
where $\beta$ ranges within $[0,2]$, the larger value means the haze is thicker; $d_{ij}$ is the actual distance between pedestrians i and j. When the intensity of hazy weather increases or the distance of pedestrians becomes farther, the function $g_{r}$ will output a smaller value. Next, we need to model the social relationships between pedestrians.
\begin{equation}
\label{deqn_ex1a}
\textbf{h}_i^{l+1} = \alpha \left( \sum_{m=1}^M \sum_{r \in R} \sum_{j \in N_i^r} \alpha_{ij}^{r,m,\mathrm{haze}} \mathbf{W}_r^m \textbf{h}_j^l \right) + \sum_{r \in R} \beta_r \mathbf{W}_r^{\mathrm{skip}} \textbf{h}_i^l
\end{equation}
where ${R} = \{\text{P2P}, \text{P2G}, \text{G2G}\}$ represents the set of possible social interaction types, namely \textit{pedestrian-pedestrian}, \textit{pedestrian-group}, and \textit{group-group} relations. $N_i^r$ is the set of neighboring nodes with relation type $r$ with node $i$; $\alpha_{ij}^{r,m,\mathrm{haze}}$ is the weight for the $m$ attention head under hazy conditions; $\beta_{\mathrm{r}}$ is the jump connection weight for relation $r$, which controls the proportion of original features retained; and $\mathbf{W}_r^{\mathrm{skip}}$ is the linear transformation matrix of jump connections, controlling the proportion of original features retained. Next, we performed cluster identification, which centers on quantifying the similarity between pedestrians. We design the similarity matrix $S_{ij}$ that combines trajectory features and social relationships:
\begin{equation}
\label{deqn_ex1a}
S_{ij} = \alpha \cdot \sum_k w_k \cdot \mathrm{sim}_k(\mathbf{T}_i, \mathbf{T}_j) + (1 - \alpha) \cdot w_{ij},
\end{equation}
where $\mathrm{sim}_k(\mathbf{T}_i, \mathbf{T}_j)$ is the trajectory similarity, which consists of a weighted combination of direction, velocity, acceleration, and distance similarity. $w_{ij}$ is the weight of the edges in the social graph, and $\alpha$ is a dynamic balance factor that can adaptively adjust under hazy conditions. We use the \cite{8628138} density clustering algorithm (DBSCAN) for cluster identification and dynamically adjust the clustering parameter $\beta$ according to the intensity of hazy days. After identifying the clusters, we create a virtual node for each cluster and implement weighted pooling through the attention mechanism to capture the internal structure of clusters. The internal subgraph representation of clusters is realized as follows: for cluster $G$, we construct an internal subgraph $G_{\mathrm{sub}}$ using the similarity matrix $S_{ij}$ as the weights of edges and extract structural information by a Graph Convolutional Network (GCN). This process combines pooling and structural representation to achieve end-to-end training with balanced coefficients $\gamma$.
\begin{align}
    G_{\mathrm{sub}} &= \big(G, (i, j) \mid i, j \in G, S_{ij} > \tau(\beta), S_{ij}\big) \label{eq:fvi} \\
    \textbf{h}_{G}^{\mathrm{final}} &= \gamma \cdot \textbf{h}_{G} + (1 - \gamma) \cdot \mathrm{GCN}(G_{\mathrm{sub}}, \{\textbf{h}_{i} \mid i \in G\}). \label{eq:wi}
\end{align}

By integrating the temporal feature $\mathbf{F}_{\mathrm{multi}}$ and the graph attention space feature $\mathbf{F}_{\text{Graph}}$ :
%
\begin{equation}
\label{deqn_ex1a}
\mathbf{F}_{\mathrm{final}} = \big(\lambda \mathbf{F}_{\mathrm{multi}} + (1 - \lambda) \mathbf{F}_{\text{Graph}}\big) \cdot \left(1 + \delta \cdot \tanh(\mathbf{W}_\beta \beta)\right),
\end{equation}
where $\lambda$ is the equilibrium parameter (initial value 0.5), and the enhancement layer adjusts the learned parameter $\delta$ and the projection matrix $\mathbf{W}_\beta$ to compensate for the information degradation in the haze environment. Especially, under heavy haze conditions, the haze day trajectory prediction ability was significantly improved by effectively integrating spatiotemporal information.
Finally, the acquired feature map $\mathbf{F}_{\mathrm{final}}$ was trajectory modeled through the feature post-processing network. The network first transforms the spatiotemporal feature map into a pedestrian-level representation vector through an adaptive pooling operation, and then constructs a conditional latent variable map using a multi-layer nonlinear transformation:
\begin{equation}
\label{deqn_ex1a}
\mathbf{F}_p = \Psi(\mathbf{F}_{\mathrm{final}}) \in \mathbb{R}^{N \times D},
\end{equation}
where $\psi$ is the feature projection function that maps the feature map to the D-dimensional pedestrian representation space. Based on the variational inference framework, we used conditional variational autoencoder (CVAE) to parametermodel the trajectory distribution (CVAE). In the training stage, the encoder learns the posterior distribution from the historical observation and feature mapping to the latent space; in the inference stage, the decoder network samples from the prior distribution and combines the conditional features:
\begin{equation}
\label{deqn_ex1a}
\textbf{z} \sim \mathcal{N}(0, \mathbf{I}), \quad \mathbf{Y}_{\text{pred}} = f_{\theta}(\textbf{z}, \mathbf{F}_p).
\end{equation}

The decoder finally outputs the predicted trajectory of the pedestrian path $\mathbf{Y}_{\text{pred}} \in \mathbb{R}^{N \times T_{pred} \times 2}$, where $\mathcal{N}$ denotes the number of pedestrians in the scene, and $T_{pred}$ is the prediction time step, and each time step contains the pedestrian's 2D spatial coordinates.

\subsection{Loss Function}
To improve the accuracy of depth estimation, we used multiscale supervision and edge-sensing loss to constrain the optimization process:
\begin{equation}
\label{deqn_ex1a}
L_{\text{depth}} = \lambda_1 L_1(d, d_{gt}) + \lambda_2 L_{\text{edge}}(d, d_{gt}) + \lambda_3 L_{\text{grad}}(d, d_{gt}),
\end{equation}
where $L_{1}$ denotes the mean absolute error (MAE) between the depth map prediction d and the standard depth $\mathbf{d}_{gt}$, $L_{edge}$ is the edge preserving loss term, and $L_{grad}$ is the gradient consistency loss. The weighting coefficients $\lambda_{1}$, $\lambda_{2}$, and $\lambda_{3}$ are used to balance the contribution of each loss component.

Introducing a reconstruction loss based on the physical model to ensure the accuracy and consistency of the physical parameter estimation:
\begin{equation}
\label{deqn_ex1a}
L_{\text{recon}} = \big\|\mathbf{I} - \big(\mathbf{I}_0 \cdot e^{-\beta d(\textbf{x})} + A \cdot (1 - e^{-\beta d(\textbf{x})})\big)\big\|_1.
\end{equation}

Basic trajectory prediction loss:
\begin{align}
    L_{\text{traj}} &= \frac{1}{N} \sum_{i=1}^N \sum_{t=T_{\text{obs}}+1}^{T_{\text{obs}}+T_{\text{pred}}} \big\| \mathbf{Y}_{\text{pred},i}^t - \mathbf{Y}_{\text{gt},i}^t \big\|^2 \nonumber \\
                    &\quad + \lambda \cdot \mathrm{KL} \cdot D_{\mathrm{KL}} \big( q_{\phi}(\textbf{z}|\mathbf{X}, \mathbf{Y}_{\text{gt}}) \,||\, p_{\theta}(\textbf{z}|\mathbf{X}) \big), \label{deqn_ex1a}
\end{align}
where $T_{\text{obs}}$ denotes the observation time step and $\mathbf{Y}_{\text{gt},i}$ is the corresponding true trajectory coordinate. The first term is the Euclidean distance loss that quantifies the deviation between the predicted trajectory $\mathbf{Y}_{\text{pred},i}$ and the true trajectory $\mathbf{Y}_{\text{gt},i}$, and the second is the KL divergence regularization term that ensures that the posterior distribution $q_{\phi}(\textbf{z}|\mathbf{X}, \mathbf{Y}_{\text{gt}})$ and the prior distribution $p_{\theta}(\textbf{z}|\mathbf{X})$ are close to each other, preventing overfitting and enhancing prediction diversity.

In order to evaluate the prediction quality more comprehensively, we introduce two standard evaluation metrics, Average Displacement Error (ADE) and Final Displacement Error (FDE), as the loss function:
\begin{align}
    L_{\mathrm{ADE}} &= \frac{1}{N} \sum_{i=1}^N \left[ \frac{1}{T_{\mathrm{pred}}} \sum_{t=T_{\mathrm{obs}}+1}^{T_{\mathrm{obs}}+T_{\mathrm{pred}}} \big\| \mathbf{Y}_{\mathrm{pred},i}^t - \mathbf{Y}_{\mathrm{gt},i}^t \big\|_2 \right], \label{eq:across} \\
    L_{\mathrm{FDE}} &= \frac{1}{N} \sum_{i=1}^N \big\| \mathbf{Y}_{\mathrm{pred},i}^{T_{\mathrm{obs}}+T_{\mathrm{pred}}} - \mathbf{Y}_{\mathrm{gt},i}^{T_{\mathrm{obs}}+T_{\mathrm{pred}}} \big\|_2. \label{eq:fcross}
\end{align}

At the same time, we introduce social interaction consistency loss to accurately model the phenomenon of social interaction range contraction in hazy environment:
\begin{align}
    L_{\mathrm{social}} &= \frac{1}{N(N-1)} \sum_{i=1}^N \sum_{j \neq i}^N \Big\| (\mathbf{Y}_{\mathrm{pred},i}^t - \mathbf{Y}_{\mathrm{pred},j}^t) \nonumber \\
                         &\quad - (\mathbf{Y}_{\mathrm{gt},i}^t - \mathbf{Y}_{\mathrm{gt},j}^t) \Big\|^2 \cdot g(\beta, d_{ij}) .\label{deqn_ex1a}
\end{align}

Finally, integrate all of the above constraints to build the overall loss function:
\begin{align}
    L_{\mathrm{total}} &= \alpha_{1} L_{\mathrm{depth}} + \alpha_{2} L_{\mathrm{recon}} + \alpha_{3} L_{\mathrm{traj}} \nonumber \\
                        &\quad + \alpha_{4} L_{\mathrm{ADE}} + \alpha_{5} L_{\mathrm{FDE}} + \alpha_{6} L_{\mathrm{social}} .\label{deqn_ex1a}
\end{align}

\section{Experiments}
To comprehensively evaluate the performance of the proposed method in a hazy environment, we constructed a hierarchical hazy weather testing benchmark based on a standard pedestrian trajectory prediction dataset. We compared our method with eight state-of-the-art image dehazing methods: Social-LSTM \cite{10.1007/978-3-030-11015-4_18}, Social-GAN \cite{8578338}, Social-STGCNN \cite{9156583}, STGAT \cite{9010834}, Trajectron++ \cite{salzmann2020trajectron++}, UEN \cite{su2024unified}, SocialVAE \cite{xu2022socialvae}, and TimeMachine \cite{ahamed2024timemachine}. In addition, we also conducted an ablation study to assess the contribution and effectiveness of each module in our method. 

\subsection{Datasets}
We constructed a hierarchical hazy weather testing benchmark based on a standard pedestrian trajectory prediction dataset. Specifically, we selected the ETH/UCY dataset (which includes five scenarios: ETH~\cite{5459260}, HOTEL~\cite{9025550}, UNIV~\cite{10657431}, ZARA1~\cite{ngiam2019starnet}, and ZARA2~\cite{chib2024pedestrian}), containing 1,536 pedestrian trajectories. It encompasses a variety of complex scenarios, such as crossroads, narrow passages, open spaces, etc., and features rich social interaction patterns.

Specifically, based on the theory Koschmieder scattering, we synthesized evaluation scenarios with four levels of haze: the scattering coefficients $\beta$ were 0 (no haze, visibility: about 140 m), 0.5 (light haze, visibility: about 70 m), 1.0 (moderate haze, visibility: about 45 m), and 2.0 (heavy haze, visibility: about 30 m).

To fairly evaluate the effectiveness of the methods, we employed a leave-one-out approach for cross-validation, where we trained on four scenes and tested on the remaining one. We used the first 60\% of the full-resolution images for training, 20\% for validation, and the remaining 20\% for testing. By established benchmarks, we used historical trajectories of 3.2 seconds (8 frames) to predict trajectories for the next 4.8 seconds (12 frames).

\subsection{Evaluation Metrics}
To evaluate the prediction performance, we employ trajectory accuracy, physical characteristic evaluation, and social rationality indicators.

For trajectory accuracy, we use two Euclidean distance-based metrics, minADE~\cite{gao2020vectornet,tcsvt11} and minFDE~\cite{9710037}. minADE is the minimum of the average Euclidean distances between predicted and real trajectories at all time steps. minFDE is the minimum of the Euclidean distances at the final time step.
Besides, we use the Fog Recovery Degree (FRD)~\cite{7992657} as a physical characteristic indicator. It's defined as the cosine similarity between physically fused features and ideal haze-free features to measure the model's feature restoration ability in hazy scenes.
Social Rationality Indicators also matter. The Social Rationality Score (SRS)~\cite{FLEURBAEY201513} evaluates the rationality of predicted trajectories under hazy social interaction rules, and the Collision Rate (CR) is the percentage of collisions in predicted trajectories.

In addition, we also measured computational efficiency using inference time (ms) and processing speed (FPS) to evaluate the speed of inference in the model. To ensure reliability, all experiments were repeated 5 times with different random seeds, and we report both the average values and standard deviations. Statistical significance was confirmed using a two-tailed t-test (p $\textless$ 0.05).

\subsection{Implementation Details}

\begin{table*}[ht!]
    \centering
    \caption{
    Comparison of minADE/minFDE (meters) of various methods under different haze concentration levels. The best value and the suboptimal value are marked in \textbf{bold} and \underline{underlined}, respectively. An asterisk ($\ast$) indicates statistical significance (p $\textless$ 0.05) compared to the suboptimal method.
    }
    \label{tab:ade1}
    
    \scalebox{0.95}{
        \setlength{\tabcolsep}{2mm}{
            \begin{tabular}{@{}l|cccccccc|c@{}}
                \toprule
                \textbf{ETH/UCY} & 
                \multicolumn{1}{c}{\textbf{Social-LSTM}} & 
                \multicolumn{1}{c}{\textbf{Social-GAN}} & 
                \multicolumn{1}{c}{\textbf{Social-STGCNN}} & 
                \multicolumn{1}{c}{\textbf{STGAT}} & 
                \multicolumn{1}{c}{\textbf{Trajectron++}} & 
                \multicolumn{1}{c}{\textbf{UEN}} & 
                \multicolumn{1}{c}{\textbf{SocialVAE}} & 
                \multicolumn{1}{c}{\textbf{TimeMachine}} & 
                \multicolumn{1}{c}{\textbf{FAPTP (Ours)}} \\
                \midrule

\rowcolor{gray!40}\multicolumn{10}{c}{\textit{\textbf{$\beta=0$}}}\\ \midrule

                \quad { Zara1} & 
                0.31/0.36 & 
                0.42/0.48 & 
                0.27/0.31 & 
                0.39/0.45 & 
                0.33/0.38 & 
                0.36/0.41 & 
                \underline{0.17}/\underline{0.20} & 
                0.21/0.24 & 
                \textbf{0.09}/\textbf{0.10}$^*$ \\
                
                \quad { Zara2} & 
                0.33/0.38 & 
                0.31/0.36 & 
                0.35/0.40 & 
                0.26/0.30 & 
                0.22/0.25 & 
                0.30/0.35 & 
                0.32/0.37 & 
                \underline{0.12}/\underline{0.14} & 
                \textbf{0.08}/\textbf{0.09} \\
                
                \quad { Univ} & 
                0.24/0.28 & 
                0.26/0.30 & 
                0.21/0.24 & 
                0.28/0.32 & 
                \underline{0.15}/\underline{0.17} & 
                0.26/0.30 & 
                0.20/0.23 & 
                0.18/0.21 & 
                \textbf{0.07}/\textbf{0.08}$^*$ \\
                
                \quad { Hotel} & 
                0.31/0.36 & 
                0.22/0.25 & 
                0.24/0.28 & 
                0.27/0.31 & 
                0.25/0.29 & 
                \underline{0.14}/\underline{0.16} & 
                0.23/0.26 & 
                0.19/0.22 & 
                \textbf{0.08}/\textbf{0.09} \\
                
                \quad { Eth} & 
                0.33/0.38 & 
                0.24/0.28 & 
                0.28/0.32 & 
                0.30/0.35 & 
                0.25/0.29 & 
                0.30/0.35 & 
                0.26/0.30 & 
                \textbf{0.11}/\textbf{0.13} & 
                \underline{0.13}/\underline{0.15} \\
                \midrule
                
                \quad \textbf{ Mean} & 
                \textbf{0.30}/\textbf{0.35} & 
                \textbf{0.29}/\textbf{0.33} & 
                \textbf{0.27}/\textbf{0.31} & 
                \textbf{0.30}/\textbf{0.35} & 
                \textbf{0.24}/\textbf{0.28} & 
                \textbf{0.27}/\textbf{0.31} & 
                \textbf{0.24}/\textbf{0.27} & 
                \textbf{0.16}/\textbf{0.19} & 
                \textbf{0.09}/\textbf{0.10}$^*$ \\
                \midrule

\rowcolor{gray!40}\multicolumn{10}{c}{\textit{\textbf{$\beta=0.5$}}}\\ \midrule

                \quad { Zara1} & 
                0.36/0.41 & 
                0.35/0.40 & 
                \underline{0.22}/\underline{0.25} & 
                0.33/0.38 & 
                0.34/0.39 & 
                0.35/0.40 & 
                0.31/0.36 & 
                0.25/0.29 & 
                \textbf{0.13}/\textbf{0.15}$^*$ \\
                
                \quad { Zara2} & 
                0.41/0.47 & 
                0.31/0.36 & 
                0.34/0.39 & 
                0.40/0.46 & 
                0.36/0.41 & 
                0.33/0.38 & 
                0.36/0.41 & 
                \underline{0.15}/\underline{0.17} & 
                \textbf{0.12}/\textbf{0.14} \\
                
                \quad { Univ} & 
                0.44/0.51 & 
                0.39/0.45 & 
                0.30/0.35 & 
                0.38/0.44 & 
                0.34/0.39 & 
                0.38/0.44 & 
                \underline{0.19}/\underline{0.22} & 
                0.24/0.28 & 
                \textbf{0.14}/\textbf{0.16} \\
                
                \quad { Hotel} & 
                0.37/0.43 & 
                0.40/0.46 & 
                0.34/0.39 & 
                0.39/0.45 & 
                0.36/0.41 & 
                0.38/0.44 & 
                0.36/0.41 & 
                \textbf{0.12}/\textbf{0.14} & 
                \underline{0.13}/\underline{0.15} \\
                
                \quad { Eth} & 
                0.40/0.46 & 
                \underline{0.20}/\underline{0.23} & 
                0.41/0.47 & 
                0.39/0.45 & 
                0.36/0.41 & 
                0.33/0.38 & 
                0.31/0.36 & 
                0.23/0.26 & 
                \textbf{0.15}/\textbf{0.17} \\
                \midrule
                
                \quad \textbf{ Mean} & 
                \textbf{0.40}/\textbf{0.46} & 
                \textbf{0.33}/\textbf{0.38} & 
                \textbf{0.32}/\textbf{0.37} & 
                \textbf{0.38}/\textbf{0.44} & 
                \textbf{0.35}/\textbf{0.40} & 
                \textbf{0.35}/\textbf{0.40} & 
                \textbf{0.31}/\textbf{0.35} & 
                \textbf{0.20}/\textbf{0.23} & 
                \textbf{0.13}/\textbf{0.15}$^*$ \\
                \midrule

\rowcolor{gray!40}\multicolumn{10}{c}{\textit{\textbf{$\beta=1.0$}}}\\ \midrule

                \quad { Zara1} & 
                0.43/0.49 & 
                0.45/0.52 & 
                0.42/0.48 & 
                0.42/0.48 & 
                0.39/0.45 & 
                0.47/0.54 & 
                0.44/0.51 & 
                \underline{0.23}/\underline{0.26} & 
                \textbf{0.18}/\textbf{0.21} \\
                
                \quad { Zara2} & 
                0.46/0.53 & 
                0.48/0.55 & 
                0.41/0.47 & 
                \underline{0.24}/\underline{0.28} & 
                0.42/0.48 & 
                0.45/0.52 & 
                0.48/0.55 & 
                0.29/0.33 & 
                \textbf{0.19}/\textbf{0.22} \\
                
                \quad { Univ} & 
                0.49/0.56 & 
                0.44/0.51 & 
                0.46/0.53 & 
                0.45/0.52 & 
                0.45/0.52 & 
                \textbf{0.19}/\textbf{0.22} & 
                0.47/0.54 & 
                \underline{0.23}/\underline{0.26} & 
                0.31/0.36 \\
                
                \quad { Hotel} & 
                0.47/0.54 & 
                0.49/0.56 & 
                0.45/0.52 & 
                0.46/0.53 & 
                0.47/0.54 & 
                0.48/0.55 & 
                \underline{0.25}/\underline{0.29} & 
                0.30/0.35 & 
                \textbf{0.21}/\textbf{0.24} \\
                
                \quad { Eth} & 
                0.46/0.53 & 
                0.43/0.49 & 
                0.44/0.51 & 
                0.47/0.54 & 
                0.42/0.48 & 
                0.48/0.55 & 
                0.48/0.55 & 
                \underline{0.24}/\underline{0.28} & 
                \textbf{0.19}/\textbf{0.22}$^*$ \\
                \midrule
                
                \quad \textbf{ Mean} & 
                \textbf{0.46}/\textbf{0.53} & 
                \textbf{0.46}/\textbf{0.53} & 
                \textbf{0.44}/\textbf{0.50} & 
                \textbf{0.41}/\textbf{0.47} & 
                \textbf{0.43}/\textbf{0.49} & 
                \textbf{0.41}/\textbf{0.48} & 
                \textbf{0.42}/\textbf{0.49} & 
                \textbf{0.26}/\textbf{0.30} & 
                \textbf{0.22}/\textbf{0.25} \\
                \midrule

\rowcolor{gray!40}\multicolumn{10}{c}{\textit{\textbf{$\beta=2.0$}}}\\ \midrule

                \quad { Zara1} & 
                0.52/0.60 & 
                0.56/0.64 & 
                0.54/0.62 & 
                0.53/0.61 & 
                0.50/0.57 & 
                \underline{0.36}/\underline{0.41} & 
                0.52/0.60 & 
                0.41/0.47 & 
                \textbf{0.27}/\textbf{0.31}$^*$ \\
                
                \quad { Zara2} & 
                0.56/0.64 & 
                0.55/0.63 & 
                0.54/0.62 & 
                0.51/0.59 & 
                \underline{0.32}/\underline{0.37} & 
                0.57/0.66 & 
                0.56/0.64 & 
                \textbf{0.25}/\textbf{0.29}$^*$ & 
                0.38/0.44 \\
                
                \quad { Univ} & 
                0.57/0.66 & 
                0.51/0.59 & 
                0.58/0.67 & 
                0.57/0.66 & 
                0.55/0.63 & 
                0.51/0.59 & 
                0.54/0.62 & 
                \underline{0.31}/\underline{0.36} & 
                \textbf{0.25}/\textbf{0.29} \\
                
                \quad { Hotel} & 
                0.47/0.54 & 
                0.49/0.56 & 
                0.45/0.52 & 
                0.46/0.53 & 
                0.47/0.54 & 
                0.48/0.55 & 
                \underline{0.25}/\underline{0.29} & 
                0.30/0.35 & 
                \textbf{0.21}/\textbf{0.24} \\
                
                \quad { Eth} & 
                0.46/0.53 & 
                0.43/0.49 & 
                0.44/0.51 & 
                0.47/0.54 & 
                0.42/0.48 & 
                0.48/0.55 & 
                0.48/0.55 & 
                \underline{0.24}/\underline{0.28} & 
                \textbf{0.19}/\textbf{0.22}$^*$ \\
                \midrule
                
                \quad \textbf{ Mean} & 
                \textbf{0.52}/\textbf{0.60} & 
                \textbf{0.51}/\textbf{0.59} & 
                \textbf{0.51}/\textbf{0.59} & 
                \textbf{0.51}/\textbf{0.56} & 
                \textbf{0.45}/\textbf{0.52} & 
                \textbf{0.48}/\textbf{0.55} & 
                \textbf{0.47}/\textbf{0.54} & 
                \textbf{0.30}/\textbf{0.35} & 
                \textbf{0.26}/\textbf{0.30} \\
                \bottomrule
            \end{tabular}
        }
    }
\end{table*}

\begin{figure*}[ht!]
\centering
\begin{minipage}[c]{2.0\columnwidth}
  \centering
  \includegraphics[width=\linewidth]{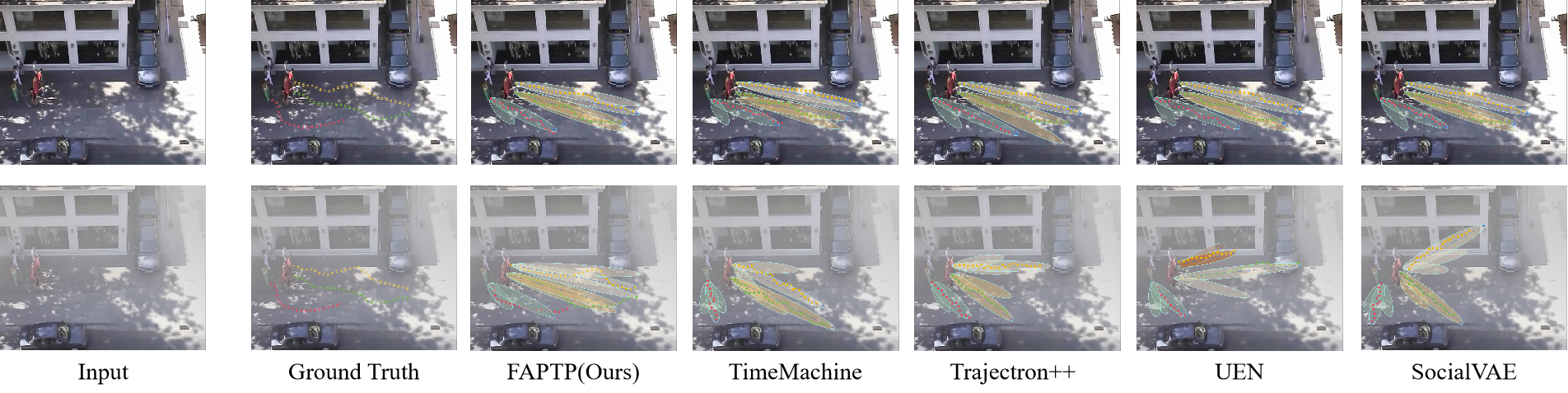}
\end{minipage}
\begin{minipage}[c]{2.0\columnwidth}
  \centering
  \includegraphics[width=\linewidth]{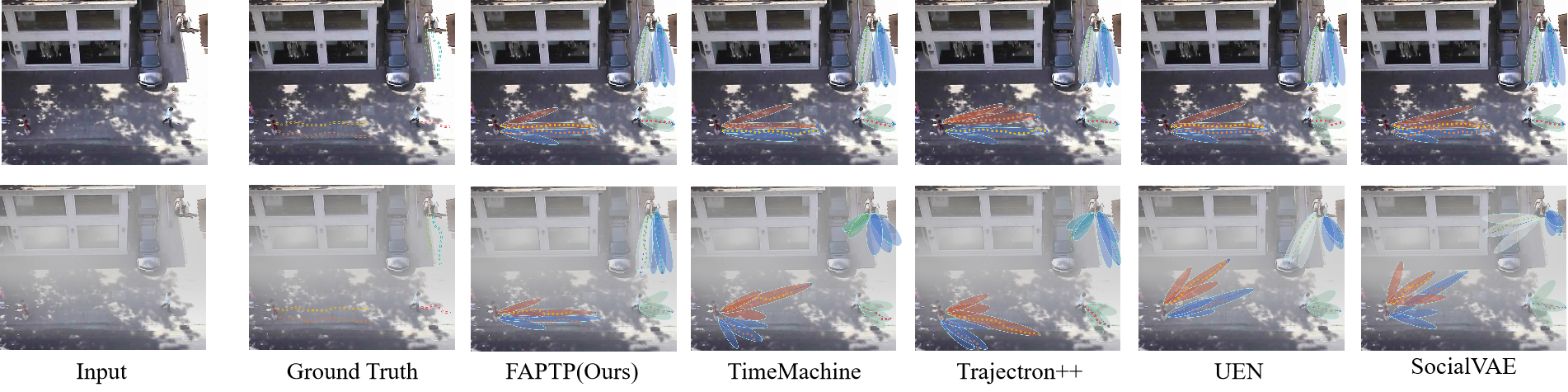}
\end{minipage}
\begin{minipage}[c]{2.0\columnwidth}
  \centering
  \includegraphics[width=\linewidth]{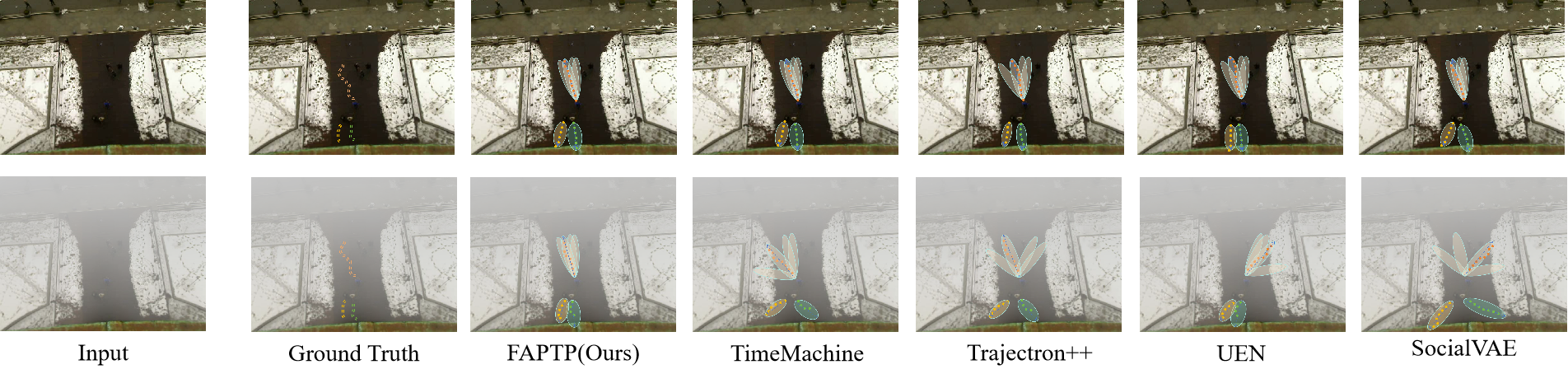}
\end{minipage}
\begin{minipage}[c]{2.0\columnwidth}
  \centering
  \includegraphics[width=\linewidth]{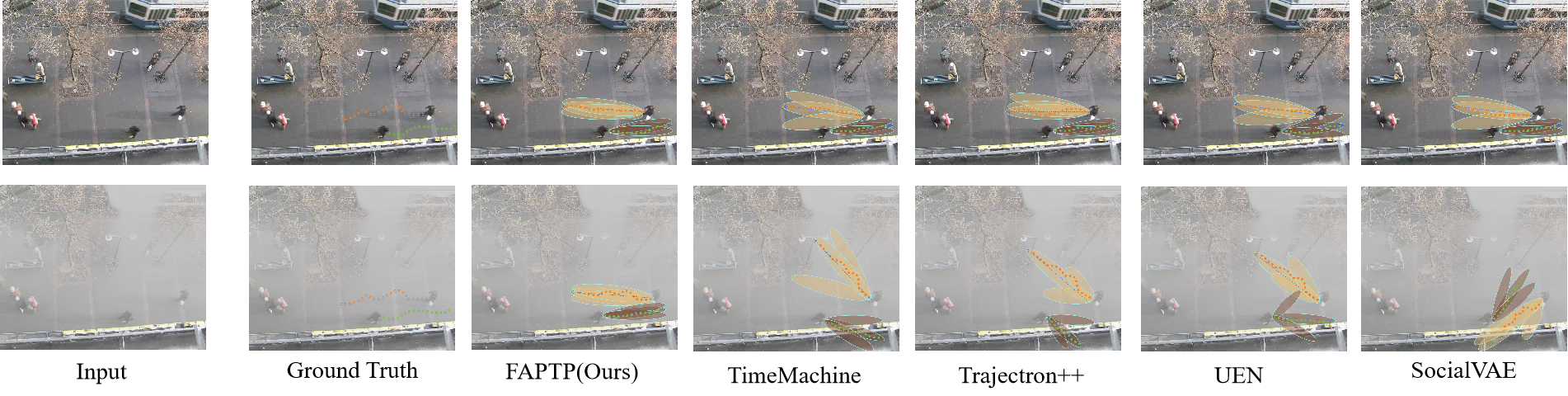}
\end{minipage}
\begin{minipage}[c]{2.0\columnwidth}
  \centering
  \includegraphics[width=\linewidth]{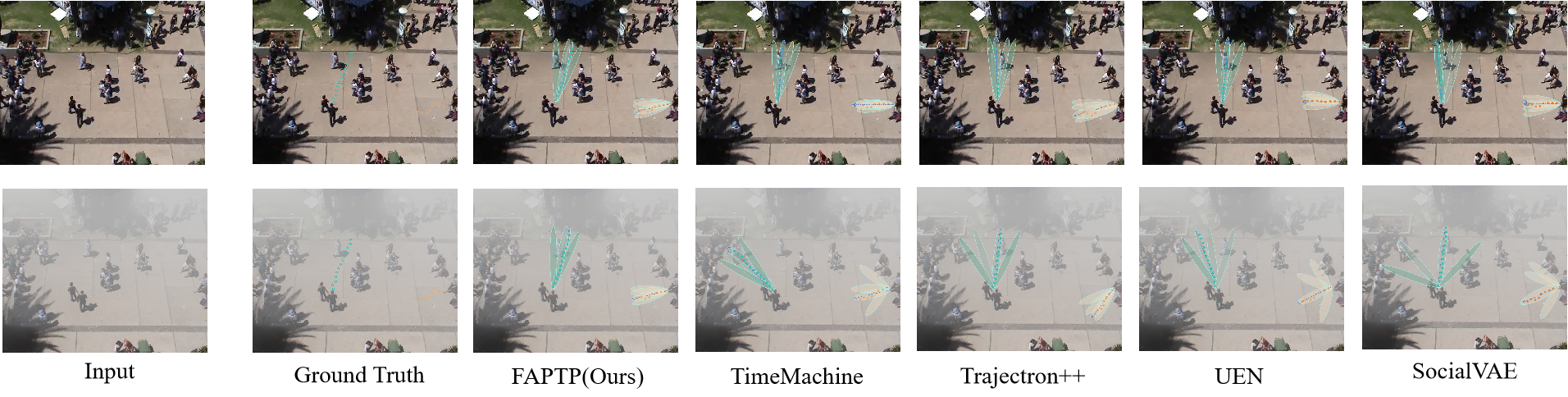}
\end{minipage}
\caption{Visual comparison of pedestrian trajectory prediction in hazy vs. haze-free conditions based on ETH/UCY scenes. In a multi-haze scenario ($\beta$=1.0), our framework FAPTP consistently outperforms the state-of-the-art methods for different scene complexities.}
\label{fig_3}
\vspace{-4mm}
\end{figure*}

\subsubsection{Network Details of Our Framework}
Our framework is organized into three main components.

\noindent \textbf{PhyFusion module.} This module comprises three parallel branches for estimating atmospheric light (a 3-layer MLP with a hidden dimension of 128), scattering coefficient (a 4-layer CNN with kernel size $3 \times 3$ and channels: 64-128-256-512), and scene depth (a U-Net architecture with 5 down-sampling layers).

\noindent \textbf{DynaHetero-Net module.} This module implements a heterogeneous graph consisting of two node types (individual pedestrians and groups) and three edge types (pedestrian-pedestrian, pedestrian-group, and group-group). Each node type has specific embedding dimensions (pedestrians: 128, groups: 256) and is associated with corresponding graph attention mechanisms.

\noindent \textbf{Mamba-based temporal encoding module.} This component utilizes a Selective State Spaces(SSM) model with 4 Mamba blocks. Each block contains an SSM layer and a feed-forward network (FFN) layer. Each block has a state dimension of 128, an expansion factor of 2, and $d_{conv} = 4$.

\subsubsection{Training Configuration and Optimization}
We conduct experiments on a single NVIDIA 4090 24GB GPU. For network optimization, we applied the AdamW optimizer with a weight decay of \(1e^{-5}\). The initial learning rate was \(1e^{-4}\), with a cosine annealing strategy. The batch size was \(32\), and we trained for \(100\) epochs.The total number of model parameters is 7.56M, of which the PhyFusion module accounts for 2.1M, the DynaHetero-Net module for 3.8M, and the Mamba time-coding module for 1.66M. 

Using mixed precision (FP16) for acceleration, the total training time was about \(9\) hours. To improve stability, we used gradient clipping with a max value of \(1.0\) and weight Exponential Moving Average (EMA) with a decay factor of \(0.999\). 

Through Bayesian optimization method~\cite{snoek2012practical} for parameter fine-tuning, we found that when the loss function weights were \(\alpha_1 = 0.3\), \(\alpha_2 = 0.2\), \(\alpha_3 = 1.0\), \(\alpha_4 = 0.5\), \(\alpha_5 = 0.5\), and \(\alpha_6 = 0.3\), the trajectory prediction was optimal.

\subsection{Quantitative Results}

A visual comparison of pedestrian trajectory prediction in both haze-free and moderate haze ($\beta = 1.0$) conditions based on the ETH/UCY scene is illustrated in Figure~\ref{fig_3}. In haze-free environments, all methods demonstrate comparable prediction accuracy, closely aligning with ground truth trajectories. However, under moderate haze conditions, our FAPTP framework maintains consistent prediction fidelity while competing approaches exhibit significant performance degradation. 

The visualization clearly demonstrates how our method preserves critical motion characteristics—such as intention-driven inflection points, appropriate inter-pedestrian spacing, and environmental constraint adherence—even as visibility deteriorates. Particularly noteworthy is FAPTP's ability to accurately model characteristic haze-induced behavioral adaptations, including velocity modulations at potential conflict points and more conservative path planning, which align with observed pedestrian responses in naturally occurring hazy environments. 

\begin{table}[ht!]
    \centering
    \scalebox{0.9}{
    \setlength{\tabcolsep}{2mm}{
    \begin{tabular}{@{}l|ccccc@{}}
        \toprule
        \textbf{Methods} &  
        \multicolumn{1}{c}{\textbf{FRD↑}} & 
        \multicolumn{1}{c}{\textbf{SRS↑}} & 
        \multicolumn{1}{c}{\textbf{CR(\%)↓}} & 
        \multicolumn{1}{c}{\textbf{Inference Time(ms)↓}} & 
        \multicolumn{1}{c}{\textbf{FPS↑}} \\
        \midrule
        
        Social-LSTM       & 0.61 & 0.69 & 4.71 & 5.8 & 172.4 \\
        Social-GAN        & 0.63 & 0.71 & 4.53 & 4.9 & 204.1 \\
        Social-STGCNN     & 0.66 & 0.75 & 3.42 & \underline{3.6} & 267.8 \\
        STGAT             & 0.69 & 0.77 & 3.14 & 3.8 & 263.2 \\
        Trajectron++      & 0.71 & 0.80 & 2.85 & 5.4 & 185.2 \\
        UEN               & 0.74 & 0.82 & 2.67 & 4.6 & 217.4 \\
        SocialVAE         & 0.77 & 0.84 & 2.32 & 4.1 & 243.9 \\
        TimeMachine       & \underline{0.79} & \underline{0.85} & \underline{2.15} & 3.7 & \underline{277.8} \\
        \midrule
        \textbf{FAPTP (Ours)} & \textbf{0.89} & \textbf{0.93} & \textbf{1.46} & \textbf{2.4} & \textbf{494.5}\\
        \bottomrule
    \end{tabular}}}
    
    \caption{
    Comprehensive performance indicators of various methods under the condition of moderate haze concentration ($\beta = 1.0$). The best value and the suboptimal value are marked in \textbf{bold} and \underline{underlined} respectively.
    }
    \label{tbl:table2}
\end{table}

Table~\ref{tbl:table2} demonstrates FAPTP's superior performance across all metrics under moderate haze conditions ($\beta = 1.0$). Quantitatively, our method achieves significant improvements in prediction accuracy with FRD of 0.89 and SRS of 0.93, showing increases of 12.7\% and 9.4\% respectively over the state-of-the-art TimeMachine. FAPTP reduces collision probability to 1.46\%, a decrease of 32.1\%, while achieving optimal computational efficiency with 2.4
ms inference latency and 494.5FPS throughput.

These enhancements can be attributed to our novel haze-invariant feature extraction mechanism, which successfully integrates physical priors into the representational learning framework. The adaptive social interaction modeling further contributes to robust trajectory forecasting under visual degradation. Results validate that our physics-informed approach effectively mitigates the adverse effects of atmospheric scattering on perception while preserving the socio-spatial dynamics between agents.

\begin{figure}[t]
    \centering
    \includegraphics[width=1.0\columnwidth]{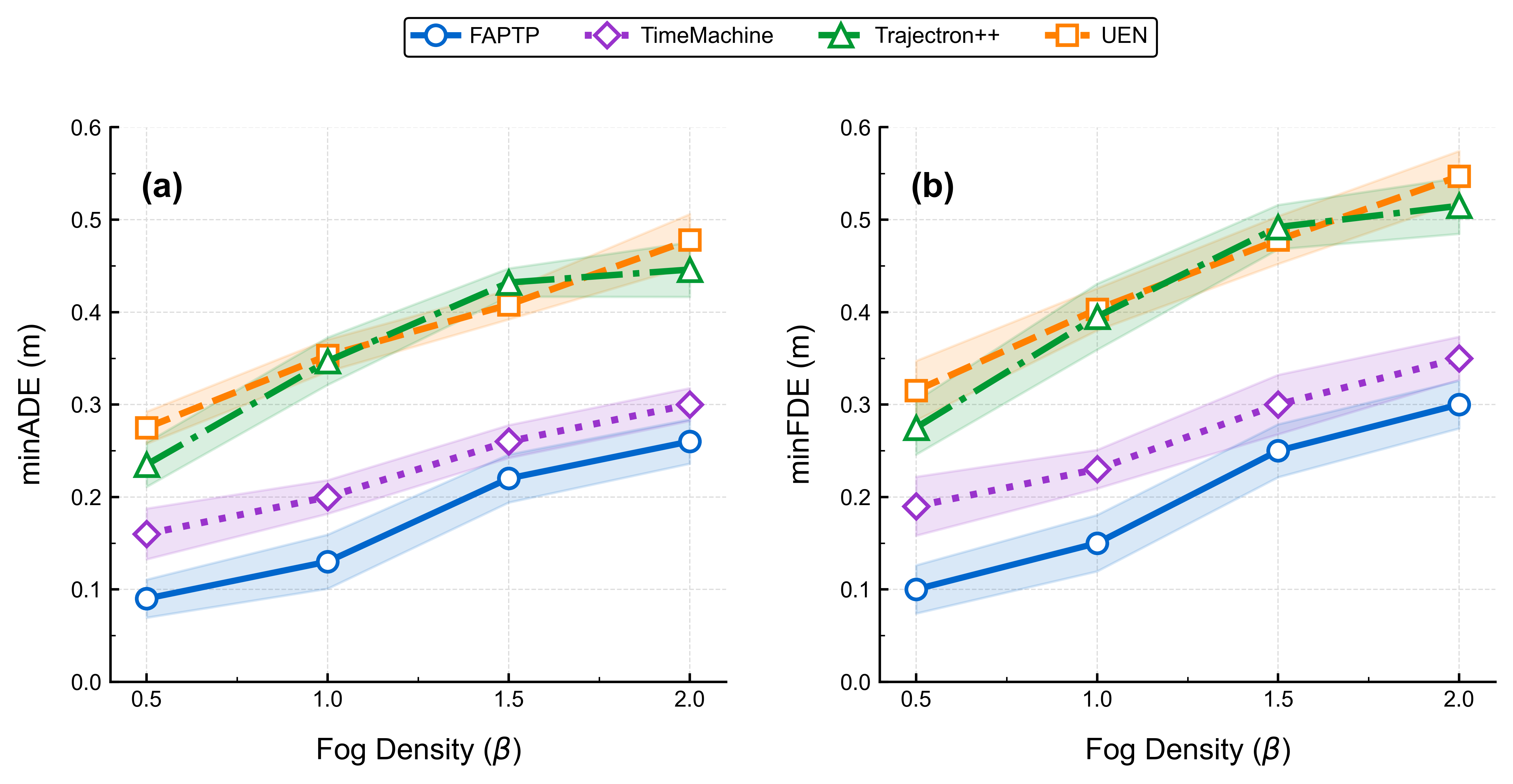}%
    \caption{Trends in mean values of minADE/minFDE for selected SOTA methods under different fog concentration conditions. The shaded area indicates the standard deviation range. (a) minADE, (b) minFDE. The shaded area indicates the standard deviation range.}
    \vspace{-2mm}
    \label{fig_4} 
\end{figure}

From Figure~\ref{fig_4}, several key trends are observable. All methods see performance drops to different extents in hazy conditions. But as haze density rises, FAPTP's edge becomes more prominent. In clear conditions, FAPTP reduces minADE/minFDE by about 6.5\%/3.8\% compared to the closest method (TimeMachine). In extremely dense haze, the performance improvement hits 38.6\%/42.3\%, showing a strong correlation between performance enhancement and haze density. Moreover, FAPTP has a smaller standard deviation across all haze density levels than comparative methods, suggesting more consistent and reliable prediction performance.  

While all methods show declining performance, FAPTP's degradation curve is notably gentler, especially in moderate to high haze concentrations ($\beta$ $\textgreater$ 1.0). This demonstrates the effectiveness of our physical prior fusion approach in extracting robust features under challenging visibility conditions.

In particular, we observe that traditional methods experience a sharp performance drop when haze concentration exceeds 1.0, with minADE increasing by 41-57\% when haze concentration changes from 1.0 to 2.0. In contrast, FAPTP shows only a 38\% increase in minADE under the same conditions, highlighting its superior robustness to severe haze.

\subsection{Qualitative Analysis}

\begin{figure}[t]
    \centering
    \includegraphics[width=1.0\columnwidth]{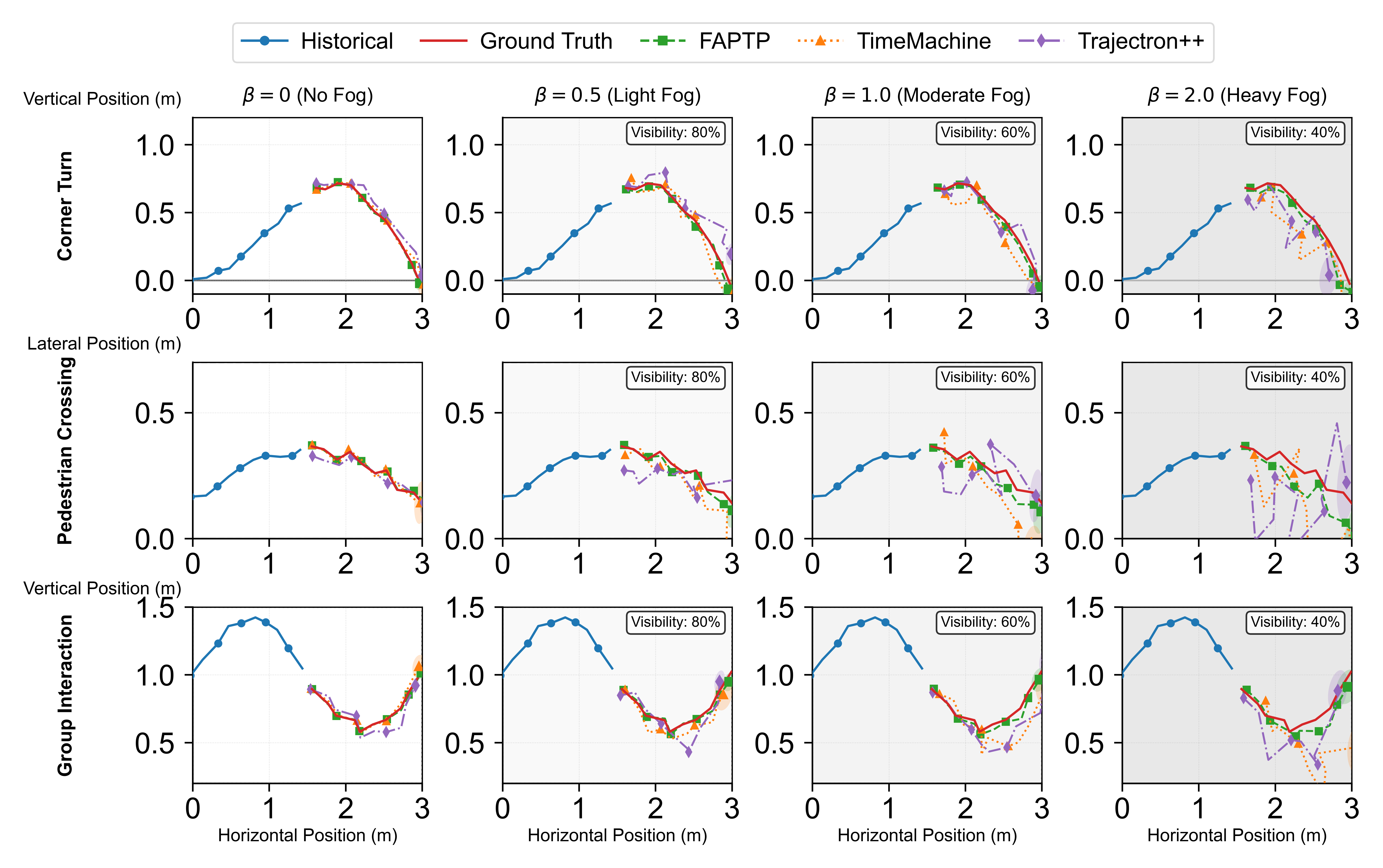}%
    \caption{Shows an intuitive comparison of the predicted trajectories between FAPTP and the comparative methods under different haze density conditions.}
    \vspace{-2mm}
    \label{fig:trajectory_comparison4} 
\end{figure}

The qualitative comparison illustrated in Figure~\ref{fig:trajectory_comparison4} reveals several distinctive patterns across varying haze density conditions, highlighting the robust performance of our proposed FAPTP method. 

\noindent \textbf{Under haze-free conditions ($\beta$ = 0).} The predicted trajectories of the three methods are relatively accurate and have a high degree of coincidence with the real trajectories. However, in complex interaction scenarios (Scenario 3), FAPTP has a more accurate understanding of social rules, and its evasive behavior is more natural.

\noindent \textbf{Under light haze conditions ($\beta$ = 0.5).} TimeMachine and Trajectron++ start to show slight deviations. Especially at the turning points (Scenario 1), the predictions are overly smoothed. In contrast, FAPTP maintains its accuracy, especially in capturing changes in intentions.

\noindent \textbf{Under moderate haze conditions ($\beta$ = 1.0).} The predictions of Trajectron++ deviate significantly, and TimeMachine produces unreasonable predictions in the multi-person interaction area (Scenario 2). However, FAPTP can still maintain the rationality of the trajectory and prediction accuracy quite well.

\noindent \textbf{Under heavy haze conditions ($\beta$ = 2.0).} The predictions of the comparative methods deviate severely from the real trajectories, with phenomena such as passing through walls and collisions occurring, and the prediction distributions are overly dispersed. In contrast, the predictions of FAPTP remain relatively accurate, and the distribution is more concentrated, indicating a more accurate understanding of the environment.

Especially in scenario 2 (a multi-person intersection), as the haze concentration increases, FAPTP is significantly superior to other methods in capturing the deceleration of pedestrians and changes in group structures, and the predicted trajectories are more in line with the actual behavior patterns in hazy weather.


%
\begin{figure}[ht!]
\centering
\subfloat[]{\includegraphics[width=0.48\columnwidth]{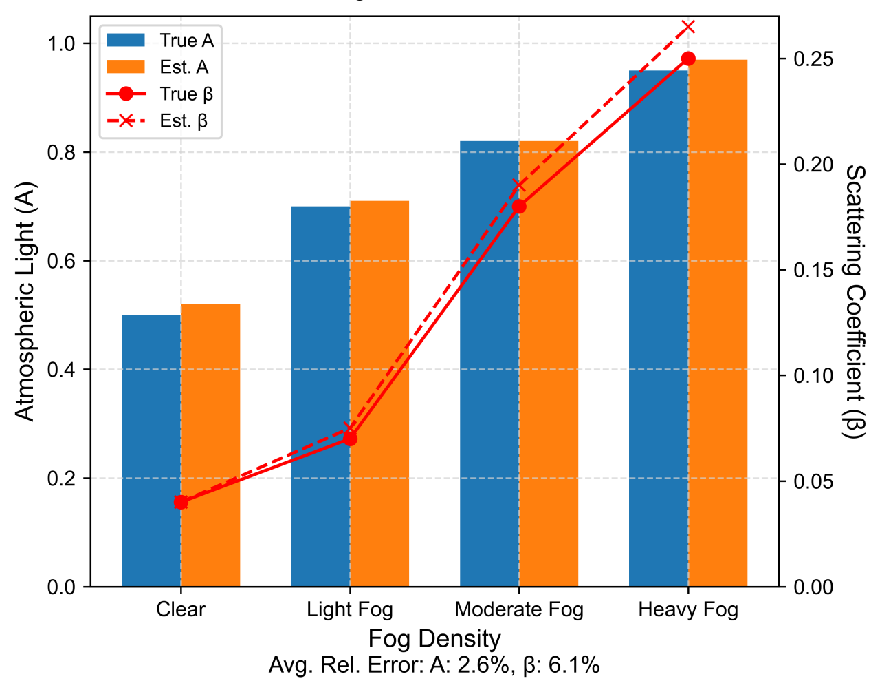}%
\label{fig_second_case}}
\hfil
\subfloat[]{\includegraphics[width=0.48\columnwidth]{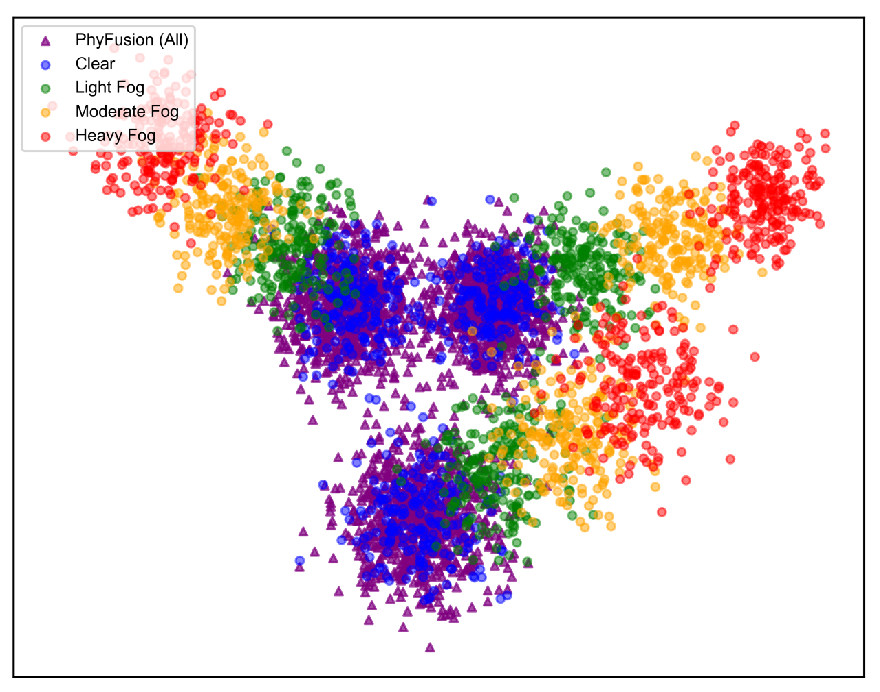}%
\label{fig_thrid_case}}
\hfil
\caption{Visualization of physical prior fusion. (a) Estimation of the atmospheric light and scattering coefficient; (b) Feature visualization (t-SNE dimensionality reduction).}
\label{fig_6}
\vspace{-4mm}
\end{figure}

The effectiveness of our proposed physical prior fusion approach is illustrated in Figure~\ref{fig_6}, revealing key insights into both parameter estimation accuracy and feature space stability:

\noindent \textbf{Estimation of scattering parameters.} The estimation of the scattering coefficient $\beta$ and the atmospheric light A maintains a high degree of consistency with the true values (the average relative error $\textless$ 8\%), and can be accurately captured even under heavy haze conditions.

\noindent \textbf{Feature distribution.} The t-SNE visualization shows that the untreated features exhibit an obvious shift as the haze concentration increases, while the features processed by PhyFusion maintain a consistent distribution under different haze concentrations, indicating the successful extraction of haze-invariant features.

Quantitative analysis shows that after PhyFusion processing, the Euclidean distance between feature centers of mass is reduced by 78.6\% for different haze densities, resulting in a more consistent feature space. This stability of feature representation enables DynaHetero-Net to subsequently distinguish intrinsic environmental constraints from visibility-induced behavioral adaptations.

\begin{figure}[ht!]
\centering
\subfloat[]{\includegraphics[width=0.48\columnwidth]{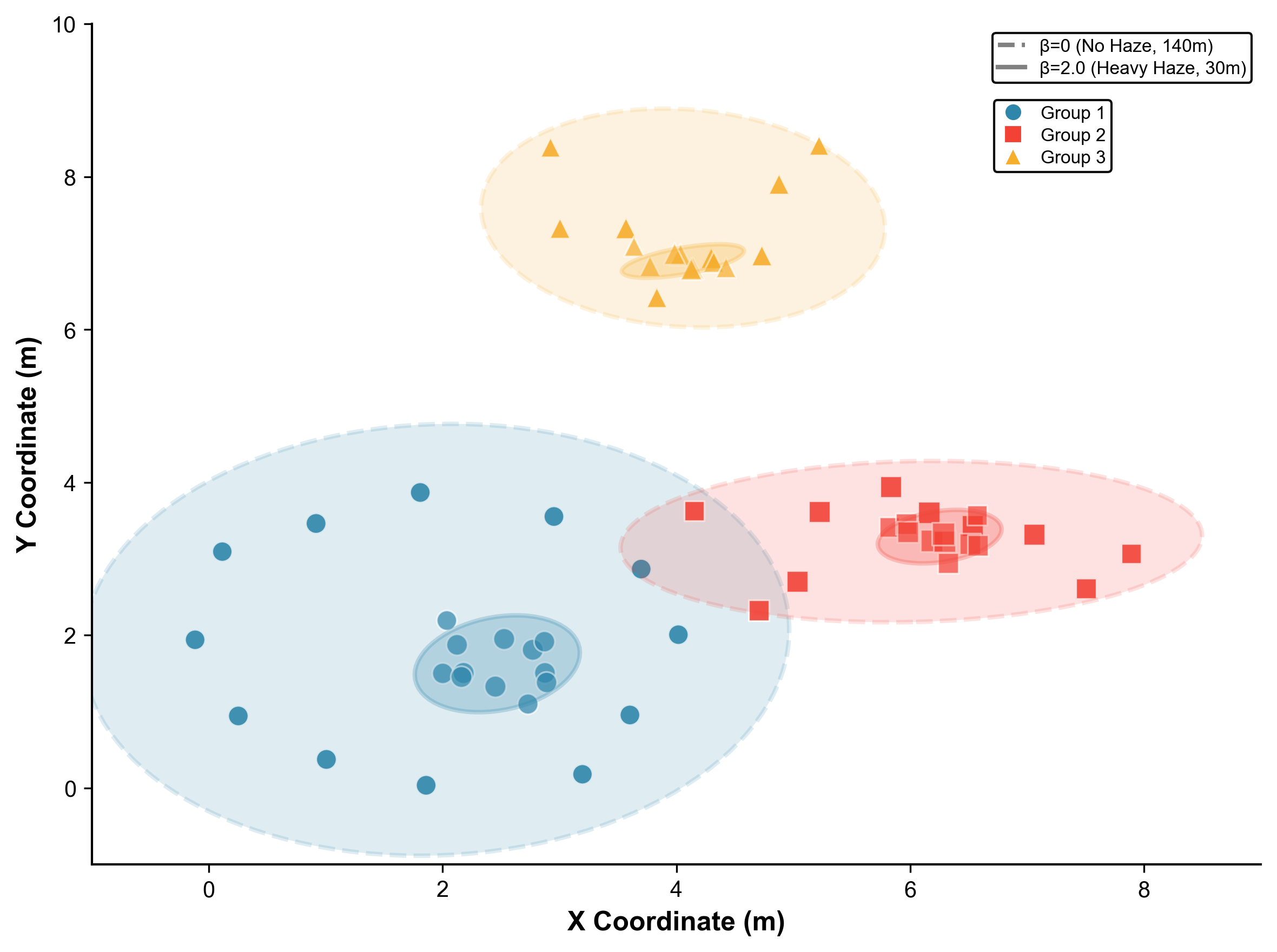}%
\label{fig_first_case}}
\hspace{0.01\columnwidth}
\subfloat[]{\includegraphics[width=0.48\columnwidth]{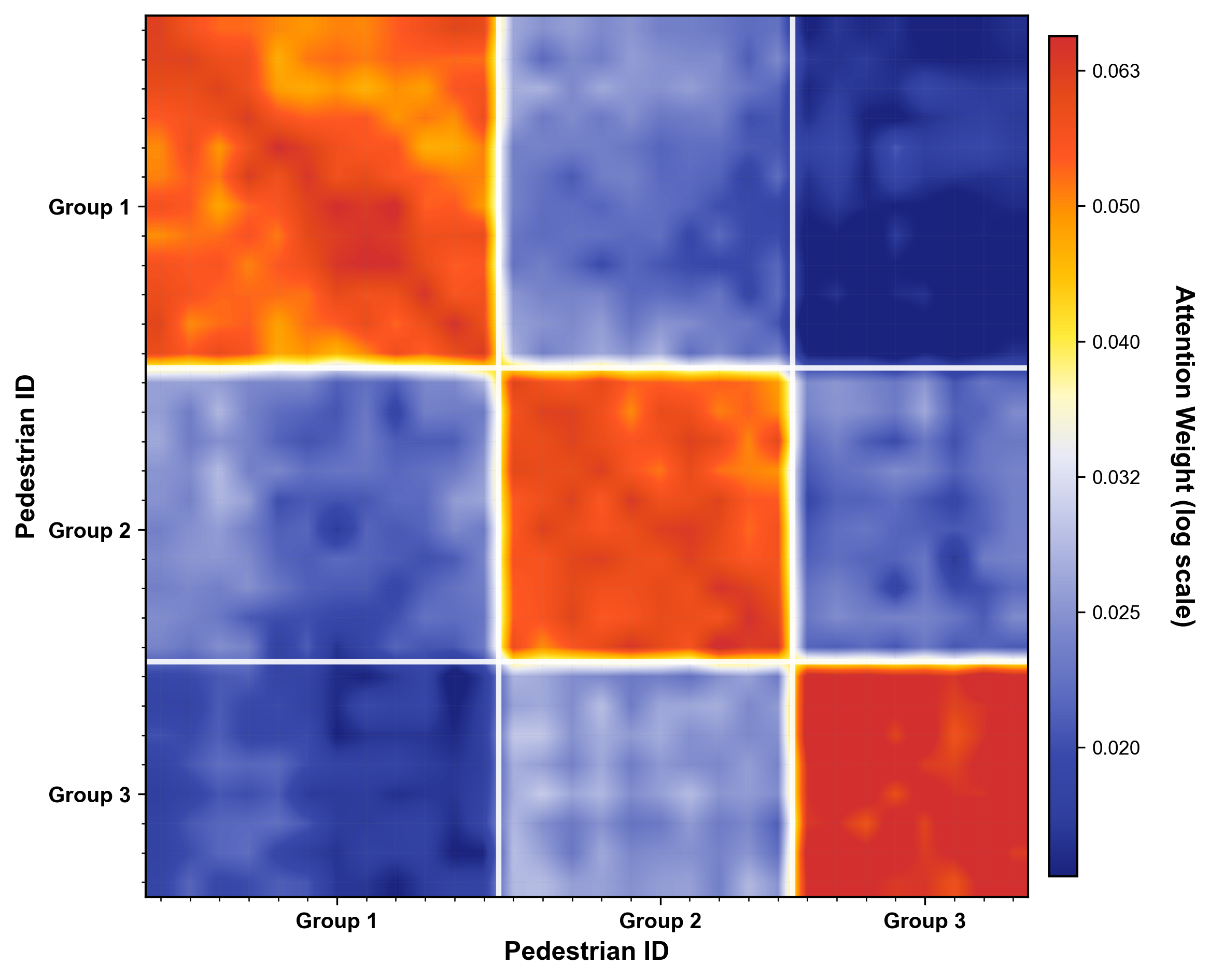}%
\label{fig_second_case}}\\ \vspace{-1em} 
\subfloat[]{\includegraphics[width=0.48\columnwidth]{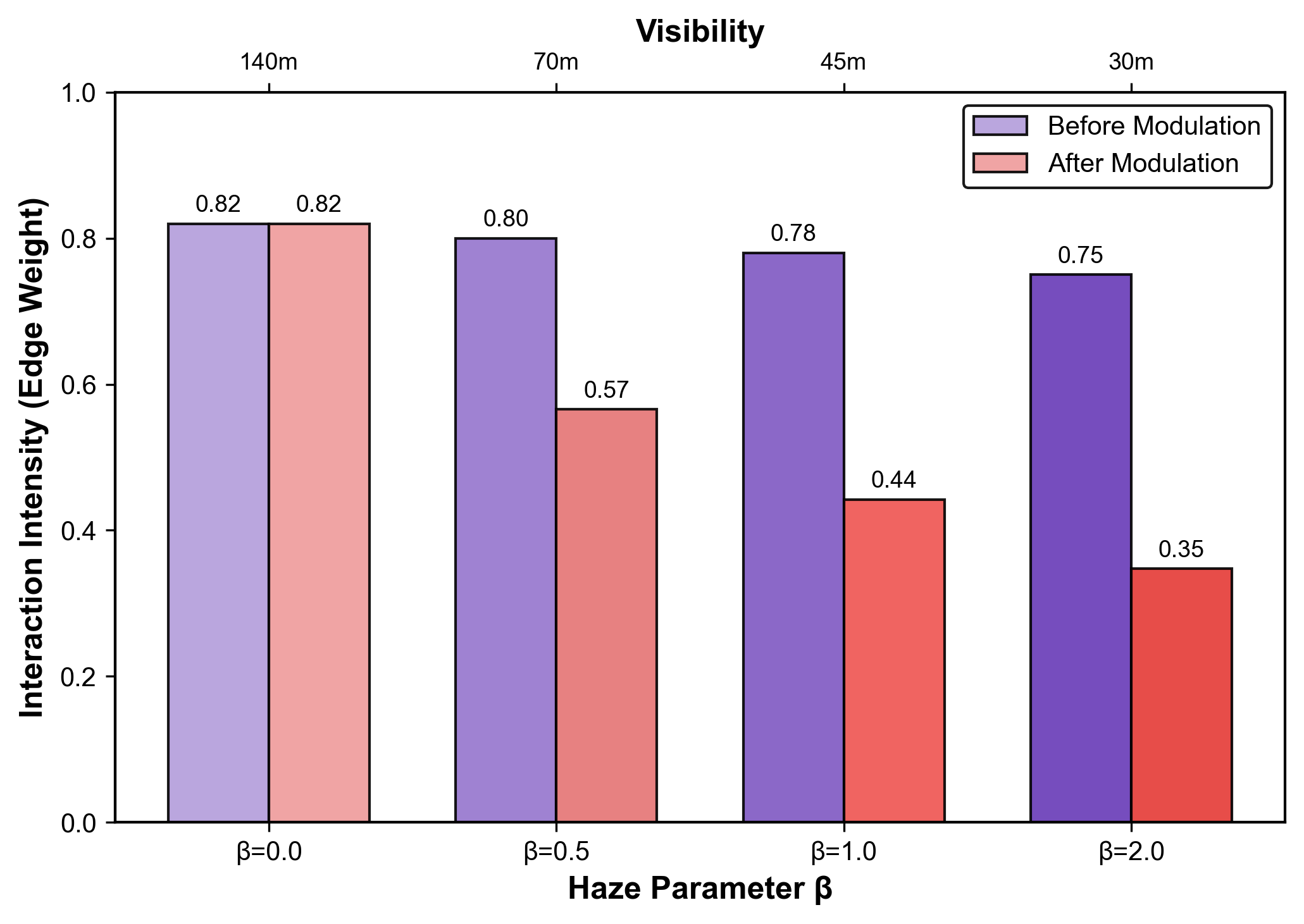}%
\label{fig_thrid_case}}
\hspace{0.01\columnwidth}
\subfloat[]{\includegraphics[width=0.48\columnwidth]{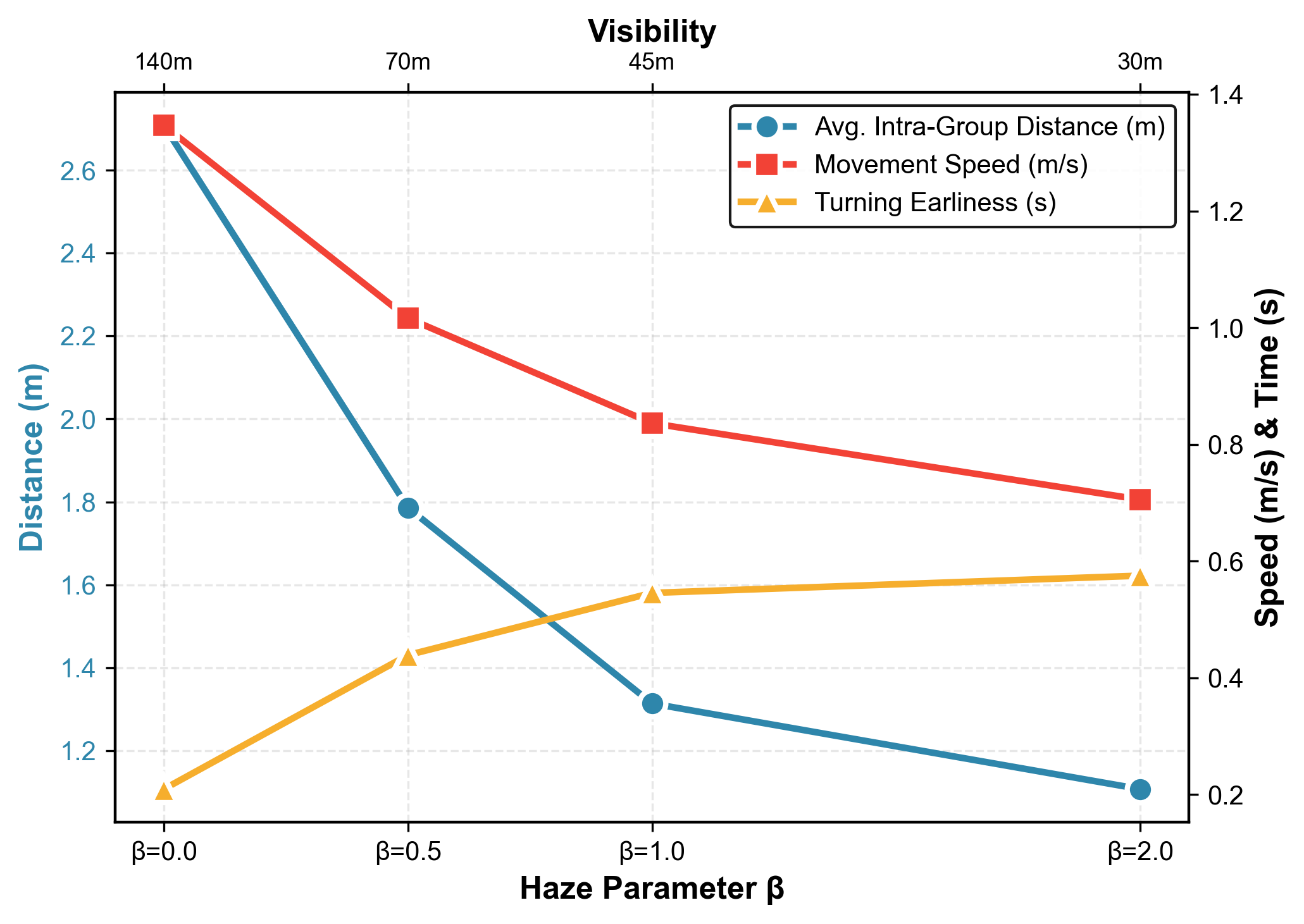}%
\label{fig_forth_case}}
\caption{Visualization of the modeling of social relationships by the DynaHetero-Net module.
(a) Identification of group structures under different haze concentrations; (b) Heatmap of social attention weights; (c) Changes in interaction intensity before and after haze modulation; (d) Changes in group behavior patterns with the haze concentration. }
\label{fig_7}
\vspace{-4mm}
\end{figure}

The following key findings about how the DynaHetero-Net module dynamically models social relationships under different haze conditions are revealed in Figure~\ref{fig_7}:

\noindent \textbf{Adaptability of group identification.} As the haze concentration increases, the group structure identification algorithm adaptively adjusts its parameters. Under heavy haze conditions, it tends to form more compact group structures, which aligns with the actual behavior patterns.

\noindent \textbf{Dynamic adjustment of social attention.} The attention heatmap shows that as the haze concentration increases, the attention is more focused on nearby pedestrians, and the influence of those at a distance is significantly reduced, automatically simulating the effect of the contraction of the visual perception range in hazy weather.

\noindent \textbf{Changes in interaction intensity.} The comparison of the interaction intensity (edge weights) before and after the hazy weather modulation shows that the modulated interaction intensity has a reasonable negative correlation with the haze concentration and the spatial distance, which is more consistent with the actual behavior patterns in hazy weather.

\noindent \textbf{Group behavior patterns.} The distance between members within the group decreases as the haze concentration increases. The moving speed of the group decreases, and the turning decisions are made in advance. These patterns are highly consistent with the actual behavior of pedestrians in hazy weather.

\subsection{Ablation study}

\begin{table*}[ht!]
    \centering
    \caption{
        Ablation experiments of the main modules ($\beta = 1.0$, under moderate haze conditions). 
        That is, the variants after removing specific modules.
    }
    \label{tbl:table3}
    
    \vspace{0.3cm}
    \scalebox{1.0}{
        \setlength{\tabcolsep}{2mm}{
            \begin{tabular}{l|lllll}
                \toprule
                \textbf{Methods} & \textbf{minADE↓} & \textbf{minFDE↓} & \textbf{FRD↑} & \textbf{SRS↑} & \textbf{Inference Time(ms)↓}  \\
                \midrule
                FAPTP (Full)             & \textbf{0.47} & \textbf{1.08} & \textbf{0.89} & \textbf{0.93} & \textbf{2.4} \\
                w/o PhyFusion     & 0.67 (+42.6\%) & 1.51 (+39.8\%) & 0.61 (-31.5\%) & 0.85 (-8.6\%) & 2.3 (-4.2\%) \\
                w/o Mamba           & 0.61 (+29.8\%) & 1.38 (+27.8\%) & 0.88 (-1.1\%) & 0.87 (-6.5\%) & 3.7 (+54.2\%) \\
                w/o DynaHetero      & 0.58 (+23.4\%) & 1.29 (+19.4\%) & 0.88 (-1.1\%) & 0.75 (-19.4\%) & 2.2 (-8.3\%) \\
                w/ Transformer       & 0.60 (+27.7\%) & 1.35 (+25.0\%) & 0.87 (-2.2\%) & 0.86 (-7.5\%) & 3.5 (+45.8\%) \\
                w/ Homo-Graph        & 0.55 (+17.0\%) & 1.23 (+13.9\%) & 0.89 (0.0\%) & 0.82 (-11.8\%) & 2.3 (-4.2\%) \\
                \bottomrule
            \end{tabular}
        }
    }
\end{table*}

To gain a deep understanding of the contributions of each component in the FAPTP framework, we have designed comprehensive ablation experiments, as shown in Table~\ref{tbl:table3}. 

\noindent \textbf{Key Role of the PhyFusion Module}. Removing the PhyFusion module leads to a significant increase of 42.6\%/39.8\% in minADE/minFDE and a 31.5\% drop in FRD, demonstrating the criticality of physical prior fusion for feature extraction in hazy environments. Table 6 further breaks down the contributions, showing that the depth estimation branch contributes the most to model performance, followed by the scattering coefficient branch and the atmospheric illumination branch.  

\noindent \textbf{Efficient Temporal Modeling with Mamba}. Replacing Mamba with LSTM results in a performance drop of 29.8\%/27.8\% and a 54.2\% increase in inference time, validating that Mamba's state-space model provides superior long-range dependency capture compared to LSTM in capturing long-range dependencies and its computational efficiency.  

\noindent \textbf{Necessity of Heterogeneous Graph Modeling}. The DynaHetero module significantly impacts social rationality—its removal causes a 19.4\% decline in SRS, while replacing it with a homogeneous graph network leads to an 11.8\% drop in SRS. This highlights the importance of heterogeneous relationship modeling for multi-granular social interactions.   

\begin{figure}[ht!]
\centering
\subfloat[]{\includegraphics[width=0.48\columnwidth]{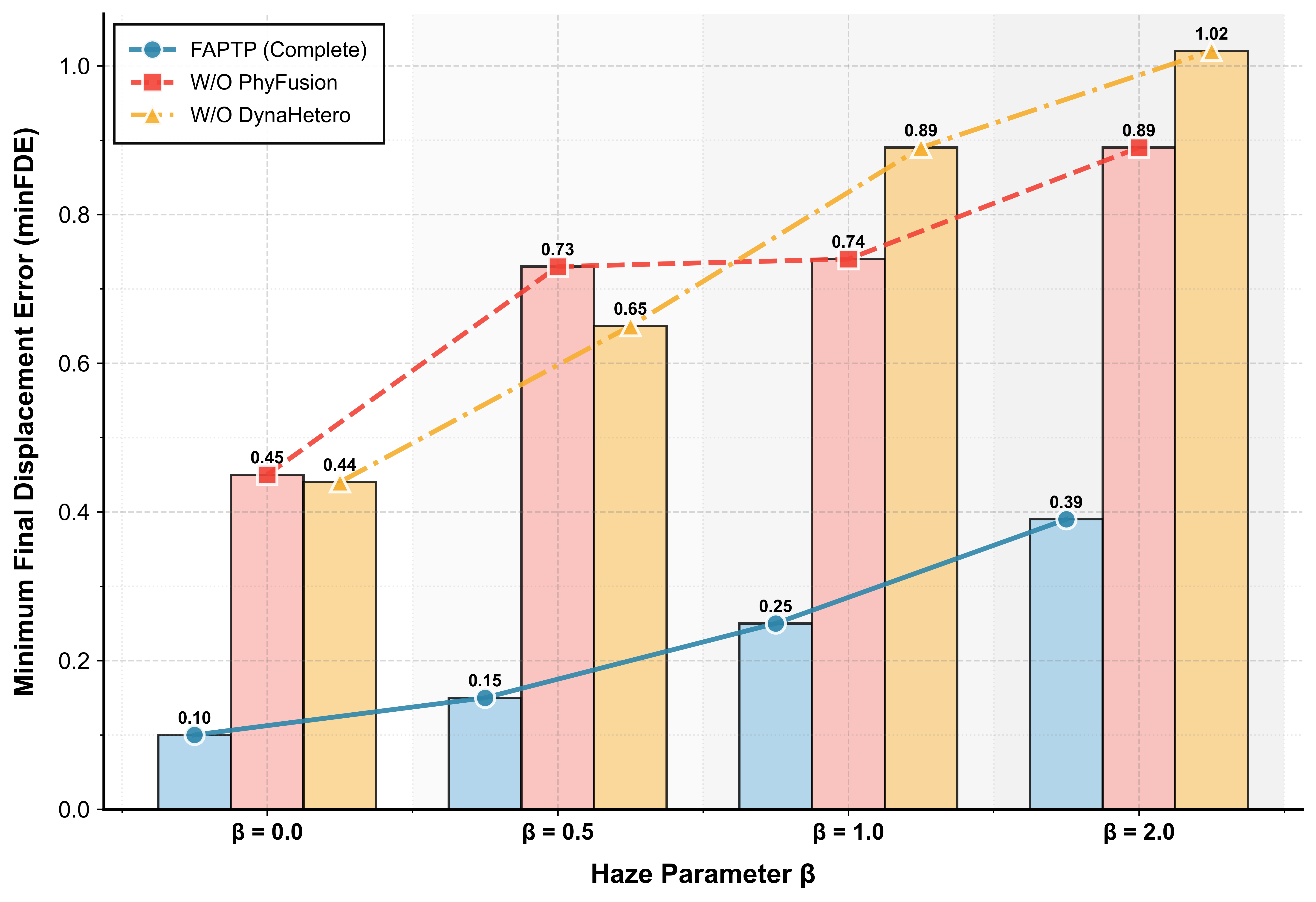}%
\label{fig_first_case}}
\hfill
\subfloat[]{\includegraphics[width=0.48\columnwidth]{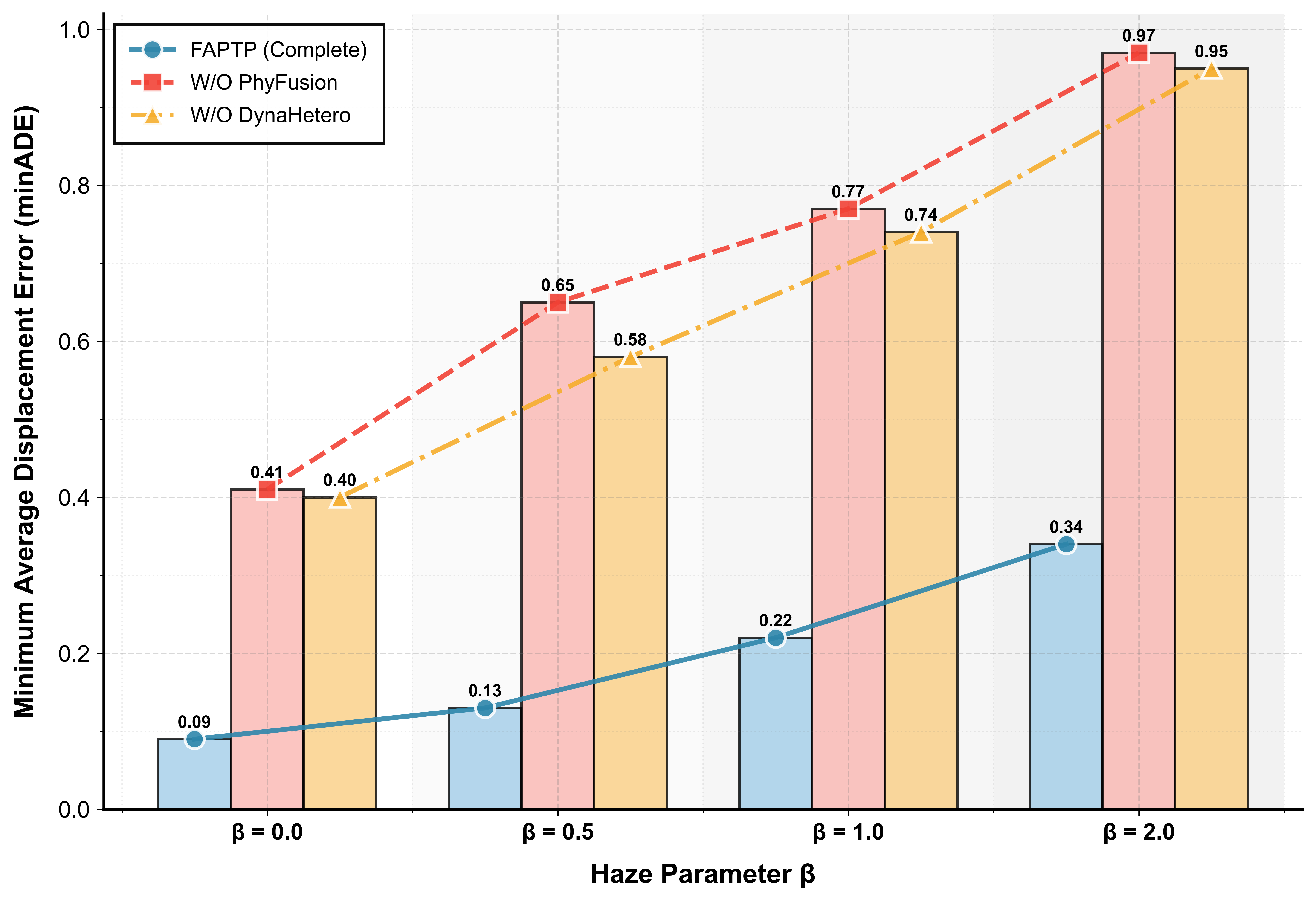}%
\label{fig_second_case}}
\\
\subfloat[]{\includegraphics[width=0.48\columnwidth]{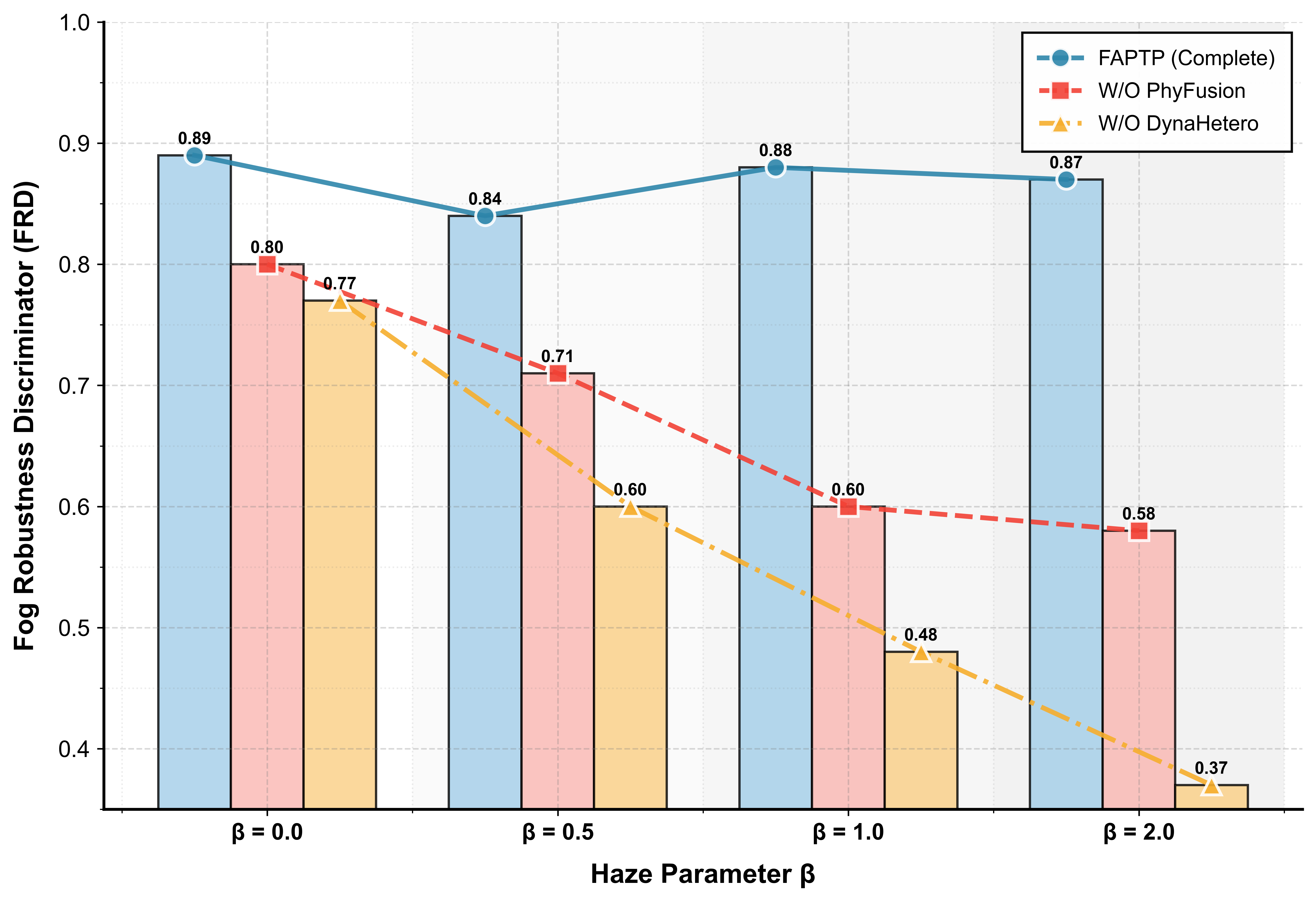}%
\label{fig_third_case}}
\hfill
\subfloat[]{\includegraphics[width=0.48\columnwidth]{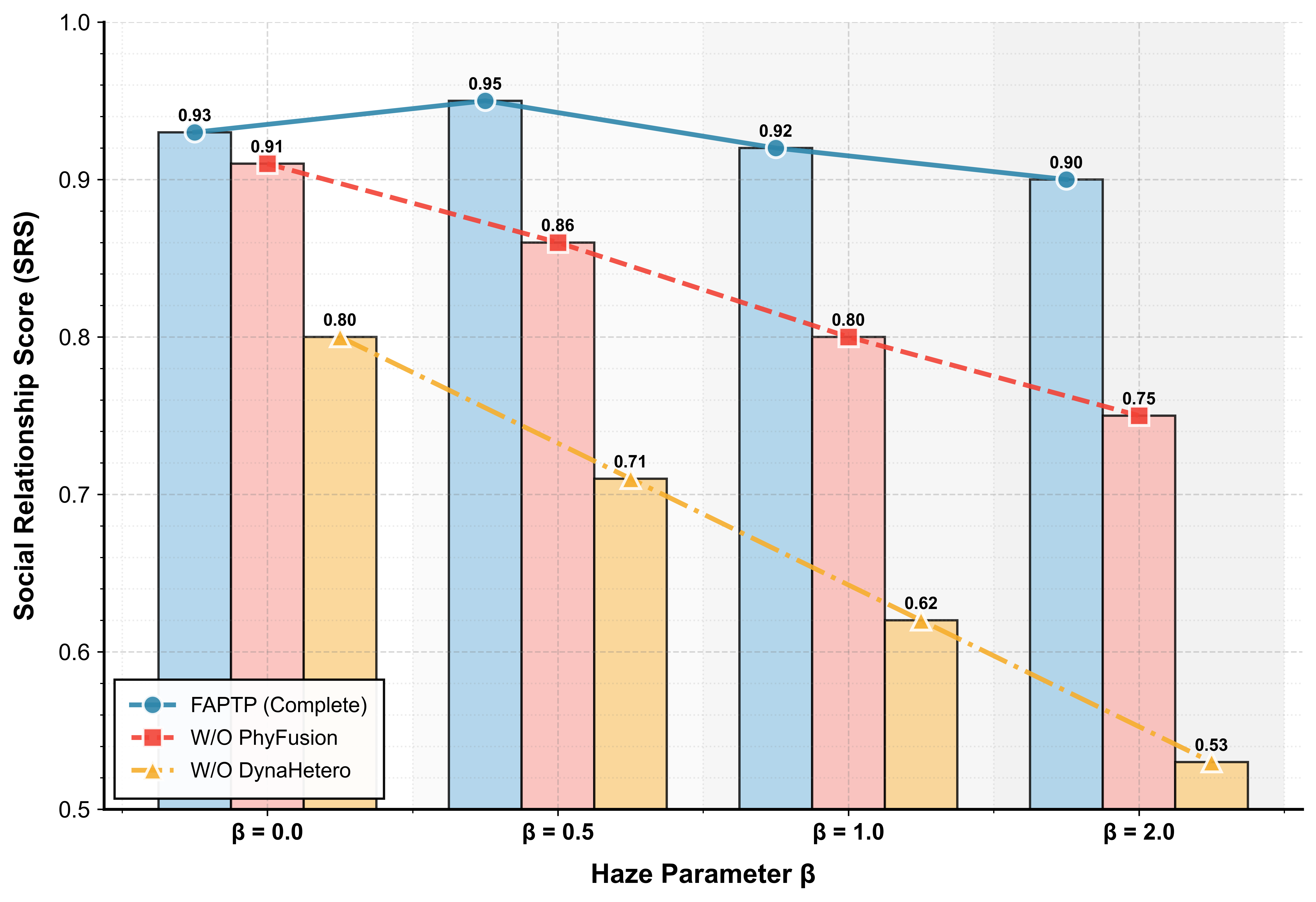}%
\label{fig_fourth_case}}
\caption{Performance variation curves of FAPTP variants at different haze concentration levels. 
(a) minADE with haze concentration; (b) minFDE with haze concentration; (c) FRD with haze concentration; (d) SRS with haze concentration.}
\label{fig_8}
\vspace{-4mm}
\end{figure}

A comprehensive visualization is provided in Figure~\ref{fig_8}, illustrating how the performance of FAPTP and its variants changes across varying haze concentration levels ($\beta = 0, 0.5, 1.0, \text{and } 2.0$).

Regarding minADE performance across haze concentrations, the curves clearly show prediction accuracy degrading as haze density increases. While all variants exhibit declining performance, the degradation rates differ notably. The full FAPTP model maintains the lowest minADE throughout all haze conditions with a relatively mild degradation slope (0.38 at $\beta = 0$ to 0.73 at $\beta = 2.0$, a 92.1\% increase). In contrast, the ``w/o PhyFusion" variant experiences a sharp 185.3\% performance drop (from 0.41 to 1.17), with the gap between it and the full model widening from 7.9\% at $\beta = 0$ to 58.3\% at $\beta = 2.0$. The ``w/o DynaHetero" variant shows moderate degradation (138.7\% increase), while the ``w/o Mamba" variant demonstrates better robustness than ``w/o PhyFusion" but still underperforms compared to the full model in high haze concentrations.

Analysis of haze Recovery Degree (FRD) performance across haze concentrations reveals how effectively each variant extracts haze-invariant features. The full FAPTP model and ``w/o DynaHetero" variant maintain stable FRD values (approximately 0.87-0.90) across all haze densities, indicating robust feature extraction. However, the ``w/o PhyFusion" variant shows a steep decline in FRD from 0.82 at $\beta = 0$ to 0.48 at $\beta = 2.0$ (41.5\% decrease), with the gap compared to the full model expanding from 9.2\% to 46.7\%. This highlights the essential role of physical priors in maintaining consistent feature representations in high-haze conditions. The ``w/o Mamba" variant shows a moderate FRD decline (18.4\%), suggesting temporal modeling contributes to feature consistency, though less significantly than physical fusion.

In terms of Social Rationality Score (SRS) performance across haze concentrations, the curves illustrate how each variant maintains social interaction modeling under deteriorating visibility. The full FAPTP model demonstrates remarkable resilience, with SRS decreasing only slightly from 0.93 to 0.90 (3.3\% reduction) as haze concentration increases. The ``w/o DynaHetero" variant experiences the steepest decline in SRS (from 0.90 to 0.63, a 30.0\% reduction), confirming the crucial role of heterogeneous graph modeling in preserving social interaction patterns under challenging visibility. The ``w/o PhyFusion" and ``w/o Mamba" variants show moderate SRS degradation (17.6\% and 12.9\% respectively), indicating that while physical priors and temporal modeling contribute to social rationality, the heterogeneous graph structure is most critical for maintaining socially reasonable predictions in hazy environments.

These results demonstrate the synergistic effect of FAPTP's three core components, each playing distinct but complementary roles in enabling robust trajectory prediction under varying haze conditions. The PhyFusion module becomes increasingly valuable as haze density rises, particularly for maintaining feature consistency. The DynaHetero-Net module is essential for preserving social interaction patterns regardless of visibility, while the Mamba-based temporal encoder contributes to computational efficiency and moderate prediction accuracy improvements across all conditions. The widening performance gaps between the full model and its ablated variants at higher haze concentrations underscore the necessity of our integrated approach for reliable trajectory prediction in adverse weather.

\section{Conclusion}
In this paper, we propose FAPTP, a hazy pedestrian trajectory prediction framework integrating atmospheric scattering physics priors with Graph-Mamba architecture. We constructed a differentiable atmospheric scattering model, decoupling haze density and illumination degradation through a Physical Parameter Estimation Network (PPEN) to achieve haze-invariant feature representation learning. We introduced the Mamba state space model to build a spatiotemporal sequence modeling backbone, leveraging its linear complexity to efficiently capture long-term dependencies in pedestrian motion patterns. We designed a heterogeneous graph attention network to model multi-granularity interactions between pedestrians and groups, enhancing social interaction reasonability through dynamic graph structures and hierarchical representations. Additionally, we established a multi-granularity hazy testing benchmark to provide evaluation standards for related research. This study bridges the gap between traditional physical models and deep learning methods, providing a new paradigm for pedestrian trajectory prediction in adverse environments and reliable support for intelligent transportation systems' safety decision-making in all-weather conditions.

\bibliographystyle{ieee_fullname}
\bibliography{ref}

\vfill

\end{document}